\documentclass{article}

\usepackage[final]{neurips_2024}

\usepackage[utf8]{inputenc} 
\usepackage[T1]{fontenc}    
\usepackage{hyperref}       
\usepackage{url}            
\usepackage{booktabs}       
\usepackage{amsfonts}       
\usepackage{nicefrac}       
\usepackage{microtype}      
\usepackage{xcolor}         
\usepackage{graphicx} 
\usepackage{amsmath}
\usepackage{amsthm}
\newtheorem{theorem}{Theorem}[section]

\newtheorem{corollary}[theorem]{Corollary}
 \newtheorem{definition}[theorem]{Definition}

\title{Counterpart Fairness – Addressing Systematic Between-group Differences in Fairness Evaluation}

\author{Yifei Wang$^{1, \dagger}$, Zhengyang Zhou$^{1, \dagger}$, Liqin Wang$^{2, 3}$, John Laurentiev$^{2}$, Peter Hou$^{2}$, \\
\textbf{Li Zhou}$^{2, 3}$, \textbf{Pengyu Hong}$^{1, *}$ \vspace{0.1cm} \\
$^{1}$Department of Computer Science, Brandeis University, Waltham, MA, USA \\
$^{2}$Brigham and Women’s Hospital, Boston, MA, USA \\ $^{3}$Harvard Medical School, Boston, MA, USA \\
$^{*}$  Corresponding author \quad
$^{\dagger}$ Equal contribution \vspace{0.1cm} \\
\texttt{\{yifeiwang, zhengyjo, hongpeng\}@brandeis.edu}, \\ 
\texttt{\{lwang, jlaurentiev, phou, lzhou\}@bwh.harvard.edu} \\
}

\begin{document}

\maketitle

\begin{abstract}
  When using machine learning to aid decision-making, it is critical to ensure that an algorithmic decision is fair and does not discriminate against specific individuals/groups, particularly those from underprivileged populations. Existing group fairness methods aim to ensure equal outcomes (such as loan approval rates) across groups delineated by protected variables like race or gender. However, in cases where systematic differences between groups play a significant role in outcomes, these methods may overlook the influence of non-protected variables that can systematically vary across groups. These confounding factors can affect fairness evaluations, making it challenging to assess whether disparities are due to discrimination or inherent differences. Therefore, we recommend a more refined and comprehensive fairness index that accounts for both the systematic differences within groups and the multifaceted, intertwined confounding effects. The proposed index evaluates fairness on counterparts (pairs of individuals who are similar with respect to the task of interest but from different groups), whose group identities cannot be distinguished algorithmically by exploring confounding factors. To identify counterparts, we developed a two-step matching method inspired by propensity score and metric learning. In addition, we introduced a counterpart-based statistical fairness index, called Counterpart Fairness (CFair), to assess the fairness of machine learning models. Empirical results on the MIMIC and COMPAS datasets indicate that
standard group-based fairness metrics may not adequately inform about the degree of unfairness
present in predictions, as revealed through CFair.
\end{abstract}

\section{Introduction}
\label{intro}
With the availability of increasingly large and complex datasets and recent advances in machine learning (ML), we are presented with unprecedented opportunities to harness big healthcare data to facilitate and optimize decision making. At the same time, the research community is increasingly acknowledging the associated difficulties in guaranteeing the precision, efficacy, and non-discrimination of ML tools deployed in real-world clinical practice. When applying ML to assist with decision-making, it is important to ensure that the algorithmic decision is fair and does not discriminate against certain groups, particularly for unprivileged populations \cite{dignum2018ethics}. This is critical because ML algorithms can perpetuate or even exacerbate existing biases if not carefully managed. In response to the need to mitigate or address ML discrimination against protected variables, computational fairness has recently emerged as an important research direction \cite{mehrabi2021survey, caton2020fairness}. Many fairness metrics have been proposed to quantify the fairness of algorithmic decisions and help train fairer ML models. 

Efforts have been devoted to research on various notions of fairness as well as how fairness is formalized in the machine learning models \cite{mehrabi2021survey, verma2018fairness, castelnovo2022clarification, gajane2017formalizing, pessach2022review}. Group fairness, one of the most popular fairness metrics, typically defines groups based on socially sensitive or legally protected variables (e.g., race, gender, age, etc.) and requires equal group-wise measures, such as equal outcomes, equal performance, equal allocation, and so on \cite{rajkomar2018ensuring}.  However, in scenarios where individuals with similar characteristics are expected to receive similar algorithmic decisions, such as in healthcare or recidivism risk predictions, group fairness can fall short. This is particularly problematic when distinct systematic differences exist between groups, making it unclear whether performance disparities reflect genuine unfairness in decisions or inherent between-group variations. Due to complex social structural inequities and the resulting significant disparities in social determinants, group-based metrics may under-appreciate the systematic between-group differences in the baseline characteristics (represented by non-protected variables) underpinning the tasks of interest. For example, differences in socioeconomic status and geographic location could lead to disparities in healthcare resource availability, disease incidence, risk factors, collecting/documenting patient records, and so on \cite{jacobs2021measurement}. This means that the baseline health characteristics of different groups can have distinct distributions. For conciseness, we refer to "systematic between-group differences" as "systematic differences" in the rest of the paper.

In reality, systematic differences could be intertwined with other issues, such as biases in collecting data (or biased sampling), to further complicate the causes of training a biased ML model \cite{suresh2019framework}. Biased sampling frequently occurs in real world applications, in which a group is not faithfully represented with respect to its true distribution in the collected data. Its effects on the distribution of the collected data can be indistinguishable from those caused by systematic differences. Hence, in this work, we treat it as part of the force creating systematic differences in data. Systematic differences would also incur confounding issues \cite{skelly2012assessing} as non-protected predictors may be associated with both protected variable(s) and outcome variable(s), which could result in training biased ML models. Therefore, we recommend a more refined and comprehensive fairness index that accounts for both the systematic differences within groups and the multifaceted, intertwined confounding effects. To this end, we make the following major contributions:
\begin{enumerate}
    \item Analyze the impact of systematic differences and biases in data collection on group fairness assessment. 
    \item Propose CFair, a novel fairness index, which evaluate fairness on counterparts comprised of pairs of similar individuals from different groups. 
    \item Develop an implementation of finding counterparts that combines propensity score matching, prior domain knowledge, and metric learning.
\end{enumerate}

We demonstrate CFair on several applications including medical treatment prediction on the MIMIC--IV (2.0) dataset \cite{mimic2022} (referred as MIMIC in the rest of paper), income prediction on the Adult dataset \cite{adult1996}, credit risk assessment on the German Banking dataset \cite{germancredit1994}, and recidivism risk predictions on the COMPAS dataset \cite{compas2016}. Empirical results on the MIMIC and COMPAS datasets illustrate that standard group-based fairness metrics may not adequately inform about the degree of unfairness present in predictions, as revealed through counterpart fairness. This insight underscores the need for CFair to capture fairness dynamics that remain hidden when relying solely on group-based metrics.

\section{Preliminaries}
Below we revisit the definitions of group fairness and examine the challenges that arise in its practical evaluation. Using the demographic parity gap, a popular group fairness metric, as a paradigmatic example, we explore how systematic differences across groups can skew fairness assessments, potentially obscuring underlying inequities. This analysis emphasizes the limitations of group fairness metrics in capturing nuanced disparities, motivating the need for of addressing systematic differences when analyzing group algorithmic fairness. Related works are discussed in Appendix~\ref{sec:related-works}.

\subsection{Revisiting Demographic Parity -- A Popular Group Fairness Index}
\label{sec:group-fairness}
Group fairness evaluates the fairness of a model across groups. In this paper, we focus on analyzing demographic parity \cite{zhao2019inherent}, one popular group fairness index \cite{narayanan2018translation}, where the binary target variable was extended to be continuous with a range of $[0, 1]$. We first considered a general decision-making system which is defined on a joint distribution $\phi$ over the triplet $T=(X, Y, Z)$, where $X \in \mathcal{X} \in \mathbb{R}^{d}$ is the input vector, $Y \in \mathcal{Y} \in [0, 1]$ is the continuous target variable, and $Z \in \{0, 1\}$ is the protected variable, e.g., race, gender, etc. We used lower case letters $x$, $y$, and $z$ to represent an instantiation of $X$, $Y$, and $Z$, respectively. To keep the notation uncluttered, for $z \in \{0, 1\}$, we took $\phi_{z}$ to denote the conditional distribution of $\phi$ given $Z=z$, and used $\phi_{z}(Y)$ to denote marginal distribution of $Y$ from a joint distribution $\phi$ over $\mathcal{Y}$ conditioned on $Z=z$.

\begin{definition}
  \textbf{(Demographic Parity)} Given a joint distribution $\phi$, a predictor $\hat{Y}$ satisfies demographic parity (DP) if $\hat{Y}$ is independent of the protected variable $Z$. 
\end{definition}

DP reduces to the requirement of $\phi_{0}(\hat{Y}=1) = \phi_{1}(\hat{Y}=1)$, if $\hat{Y}$ is a binary classifier, i.e., $\hat{Y} \in \{0, 1\}$. The reduced case indicates the positive outcome is given to the two groups at the same rate. When exact equality does not hold, we use the absolute difference between them as an approximate measure, i.e., the DP gap which is defined below.

\begin{definition}
    (\textbf{DP gap}) Given a joint distribution, the DP gap of a predictor $\hat{Y}$ in terms of $z$ is 
\begin{equation}
    \Delta_{\text{DP}}(\hat{Y}) = |\mathbb{E}[\phi_{0}(\hat{Y})]-\mathbb{E}[\phi_{1}(\hat{Y})]|
    \label{equ:dp-gap-standard}
\end{equation}
\end{definition}

For the reduced case where $\hat{Y}$ is a binary classifier, there is one equivalent expression:
\begin{equation}
    \Delta_{\text{DP}}(\hat{Y}):=|\phi_{0}(\hat{Y}=1)-\phi_{1}(\hat{Y}=1)|
\end{equation}

Pursuing algorithmic group fairness in terms of demography parity can be attempted by minimizing the DP gap. It is often impossible to have the underlying distribution $\phi$ over $(X, Y, Z)$, and thus both DP and $\Delta_{\text{DP}}$ are estimated from a given dataset. For example, suppose there are $N_0$ samples from Group $G_0$ and $N_1$ samples from Group $G_1$, and assume a function $f$ maps $X$ to $\hat{Y}$. Without loss of generality, we assume that $G_0$ represents the minority group, such that $N_0 \le N_1$ always holds. The estimation of DP gap follows 
\begin{equation}
\widehat{\Delta_{\text{DP}}}(G_0, G_1) = |\frac{1}{N_0}\sum_{x\in G_{0}}f(x)-\frac{1}{N_1}\sum_{x\in G_{1}}f(x)|
\label{equ:dp-gap-estimation}
\end{equation}

\subsection{DP Gap Distorted by Systematic Differences}
\label{sec:factors-influence-dp-gap}
Systematic differences in data, whether due to underlying group disparities or data collection biases (e.g., biased sampling), can lead to observable gaps in demographic parity (DP), as shown in Fig~\ref{fig:group-fair-fig-bias} (more discussions about the effects of biased sampling are provided in Appendix~\ref{appendix:biased-sampling}).  When data distributions between two groups differ significantly, certain individuals in one group may have characteristics that make them distinct from and non-comparable to those in another group. However, group fairness metrics \cite{dwork2012fairness, binns2020apparent} require to include all non-comparable individuals. As a result, enforcing group fairness uniformly across groups may not be as effective as expected and may even lead to unintended consequences, such as disparate impacts of models across groups \cite{fu2022fair}.

\paragraph{Remark 1.} Systematic differences can be observed in real-world applications. Fig~\ref{fig:german-feature-difference} in the Appendix demonstrates one example using the German Banking dataset \cite{germancredit1994}. If such differences are not appropriately handled during the process of fairness evaluation, incomparable samples from different groups will be compared in computing fairness metrics. The results could mislead stake holders to make inappropriate decisions that may broadly impact our society. Additionally, systematic differences could cause confounding issues and allow ML models to implicitly utilize protected information in making prediction. This motivates us to introduce a novel fairness metric in Section~\ref{sec:method-all}, which takes a data-driven approach to mitigate confounding issue and identify comparable samples from different groups for fairness evaluation.

\paragraph{Remark 2.} Confounding refers to a situation in which the effect of one variable on an outcome is mixed or obscured by the influence of another variable. Confounding can occur when there exist systematic differences between the groups being compared, allowing protected variables to be inferred from non-protected variables. The importance of considering confounding in clinical trials is well discussed in \cite{skelly2012assessing}. Confounding variables would also cause training machine learning models to amplify bias. For example, Wang et al.~\cite{wang2019balanced} showed that a trained ML model would significantly amplify biases if confounding is not addressed. Users should carefully address confounding issues when training models and evaluate their fairness.

\begin{figure}[t]
    \centering
\includegraphics[width=0.9\textwidth]{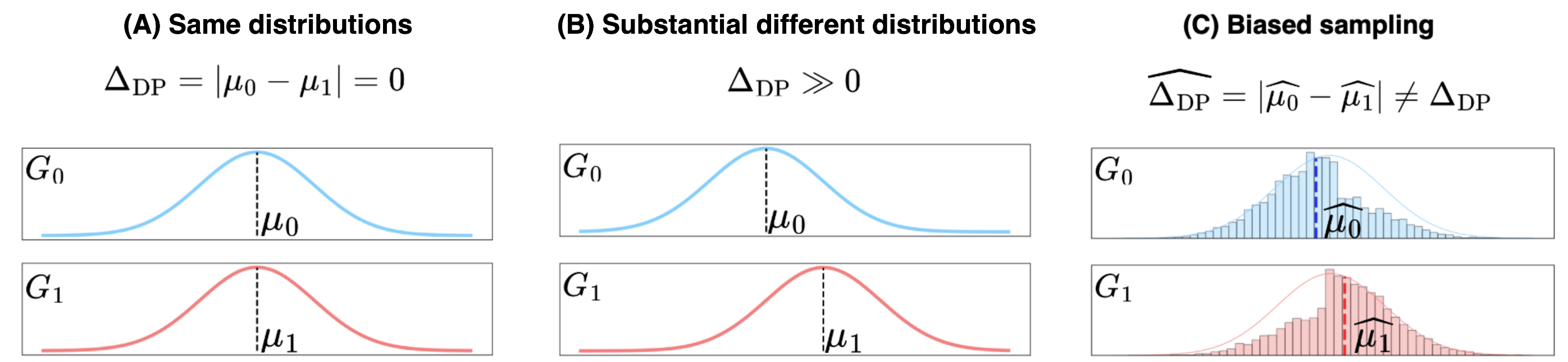}
    \caption{\small{DP gap and biases. (A) $\Delta_{\text{DP}}$ = 0 if two sample groups follow the same underlying distributions. (B) When distributions of two groups are substantially different, the true $\Delta_{\text{DP}}$ should significantly deviate from 0. (C) Biased sampling could distort DP gap estimation. In this example, the distributions (curves) of two groups are the same, and their true $\Delta_{\text{DP}}$ should be 0. However, the difference in their sample distributions (bars) leads to a large estimated $\widehat{\Delta_{\text{DP}}}$.}}
    \label{fig:group-fair-fig-bias}
\end{figure}

\section{Method}
\label{sec:method-all}
\subsection{Counterpart Fairness}
To mitigate systematic differences, we propose a novel fairness evaluation method that uses comparable samples, termed counterparts, from different groups. In this section, we begin by introducing the concept of Counterpart Fairness (CFair), followed by implementing a data-driven approach for performing CFair analysis.

\subsubsection{Counterparts}
\label{sec:counterpart-def}
We assume that the sample space $\mathcal{X}$ is equipped with a distance measurement $d(\cdot, \cdot)$, which can be designed using domain knowledge or be learned from data. Without losing generalizability, we consider two mutually disjoint groups $G_0 \subset \mathcal{X}$ and $G_1 \subset \mathcal{X}$, where $G_0$ is the protected group that is usually much smaller than $G_1$.

\begin{definition}[$\delta$-element and $\delta$-counterpart]
\label{def:counterpart-element}
     Given a threshold $\delta > 0$, one element $x \in G_0$ is a $\delta$-element, if \ $\exists x^{\prime} \in G_1$, s.t., $d(x, x^{\prime}) \le \delta$. We define $x^{\prime}$ as the $\delta$-counterpart of $x$. The definition of $\delta$-counterpart is bidirectional, that is, $x$ is also a $\delta$-counterpart of $x^{\prime}$.
\end{definition}

\begin{definition}[$\delta$-group]
\label{def:counterpart-subgroup}
     Given a threshold $\delta > 0$, let $C_{0, \delta}$ be the $\delta$-group of $G_0$, which contains all $\delta$-elements in $G_0$. Similarity, let $C_{1, \delta}$ be the $\delta$-group of $G_1$ containing all $\delta$-counterparts of the elements in $C_{0, \delta}$. $C_{0, \delta}$ and $C_{1, \delta}$ are counterpart groups of each other. 
\end{definition}


The following corollary asserts the uniqueness of $\delta$-groups. It indicates that, given a similarity function and a similarity threshold $\delta$, the chosen counterparts are consistent, which ensures stable CFair evaluation. 

\begin{corollary}
\label{thm:counterpart-unique}
    Given two groups $G_0$ and $G_1$, both $C_{0, \delta}$ and $C_{1, \delta}$ are unique.
\end{corollary} 

The proof is provided in Appendix~\ref{sec:appendix-CFair}. Note that each $\delta$-element in $C_{0, \delta}$ might have multiple $\delta$-counterparts in $C_{1, \delta}$. To avoid potential undesirable effects of imbalanced samples on fairness evaluation, we decide to choose one counterpart for each $\delta$-element in $C_{0, \delta}$, establishing 1-1 $\delta$-counterpart group relationship between $C_{0, \delta}$ and $C_{1, \delta}$.




\begin{definition}[1-1 $\delta$-counterpart groups] Let $C_{0, \delta}=\{x_{0,1}, x_{0,2}, ..., x_{0,N}\}$. For each $x_{0,i}$, one of its $\delta$-counterparts is chosen from $C_{1, \delta}$\footnote{Selection method can be either random sampling or using a deterministic method, while in this work we use a deterministic one.}, denoted as $x_{1,i}^*$. Then the subset $C_{1, \delta}^{*}:=\{x_{1,1}^*, x_{1,2}^*, ..., x_{1,N}^*\}$ is denoted as the 1-1 counterpart group of $C_{0, \delta}$. $C_{0, \delta}$ and $C_{1, \delta}^{*}$ are the 1-1 $\delta$-counterpart groups to each other.
\label{def:1-1-counterpart}
\end{definition}

With slight abuse of notations, if it can be easily determined from the context, we use "counterparts" and "$\delta$-counterpart groups" exchangeably in the rest of the paper. 


\subsubsection{CFair: Fairness on Counterparts Between Groups}
\label{sec:cfair}
CFair measures whether a model is fair across the matched counterparts. It is intuitive to extend conventional group fairness indexes for CFair analysis. For example, we can extend demographic parity (DP) to derive \textbf{counterpart DP (CDP) gap} that is defined on 1-1 $\delta$-counterparts between two groups $G_0$ and $G_1$: $\widehat{\Delta_{\text{CDP}}^{\delta}}(G_0, G_1) := \widehat{\Delta_{\text{DP}}}(C_{0, \delta}, C_{1, \delta}^{*})$. It will be shown later that CFair can be generalized to other group fairness measurements (e.g., Equal Opportunity \cite{hardt2016equality} and Sufficiency \cite{castelnovo2022clarification}) in a similar way.

In reality, it can be challenging, if not impossible, to achieve perfect fairness. A more practical question is how to evaluate the statistical significance of a disparity value measured by whatever fairness metric deployed. It should be noted that a small disparity value might still indicate unfairness. For example, if an ML model consistently exhibits subtle biases (within legally permissible bounds) favoring individuals from one group over others, it could generate a small overall disparity value, yet this still signifies a systemic bias within the model. Nevertheless, current fairness evaluation methods lack a mechanism to address this problem. CFair tackles this by offering a more rigorous fairness analysis through statistical testing. This currently is accomplished by employing the paired samples $t$-test \cite{hsu2014paired, woolson2007wilcoxon} to compare the predictions of the ML model on the 1-1 counterparts under the null hypothesis stating that there is no significant differences between the means of two paired groups. A lower $p$-value in a paired $t$-test indicates greater confidence in an ML model being systematically biased. This is one of the advantages using CFair over existing fairness indexes as they do not offer means to detect subtle but consistent bias.

One straightforward way to find 1-1 counterparts is to use propensity score matching (PSM) \cite{caliendo2008some, austin2011introduction, zhang2019balance, kline2022psmpy}. PSM is a technique commonly used to reduce bias when estimating the effect of a treatment/intervention/exposure in observational studies~\cite{hong2008effects, ye2009using, wyse2008assessing, staff2008teenage}. It aims to mimic a randomized controlled trial by creating comparable groups of treated and untreated subjects based on their propensity scores. The propensity score is calculated as the probability of receiving the treatment given a set of covariates (i.e., predictor variables). After calculating these scores, individuals in the treatment group are matched with individuals in the control group who have similar scores, balancing observed covariates across the groups to reduce confounding bias. However, since propensity scores are scalars, matched individuals with similar scores may still differ in other baseline characteristics represented by non-protected variables. To improve matching outcomes, a refined matching algorithm will be introduced in the next section.



\subsection{An Implementation of CFair}
\label{sec:cfair-method}
\begin{figure}[t]
    \centering
    \includegraphics[width=1.0\textwidth]{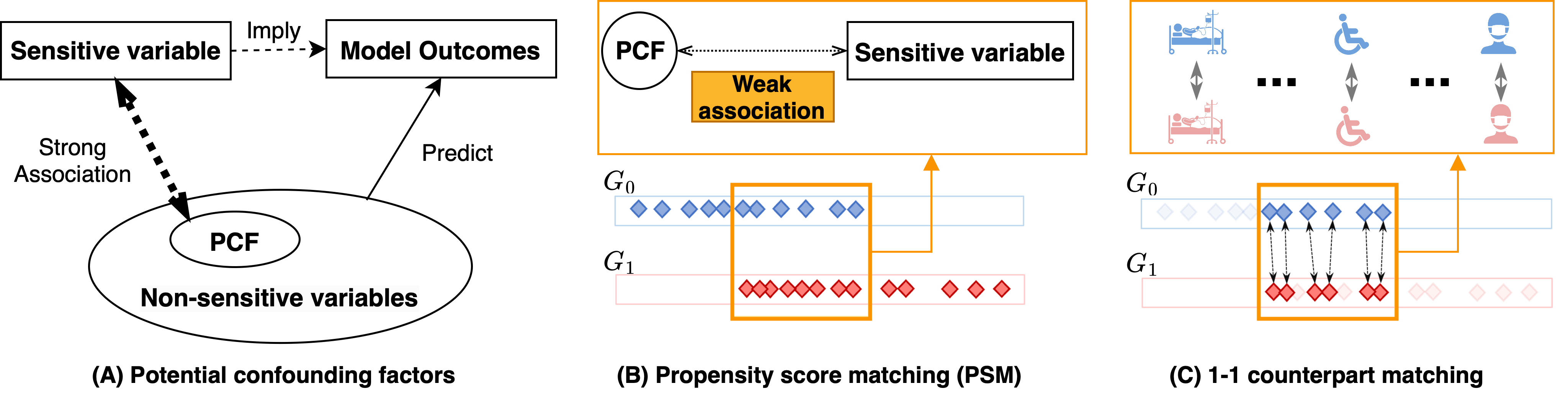}
    \caption{\small{Identify 1-1 counterparts. (A) Potential confounding factors (PCF) are a subset of non-protected variables used by an ML model for predicting outcomes, and are strongly associated with the protected variable, which can be explored by ML to accurately predict the protected variable. In this way, the protected variable can "dictate" the outcomes of the ML model (i.e., the model is biased) even though it is not used in training, which can mislead fairness evaluation. (B) Propensity score matching (PSM) is used to identify initial matches between individuals in groups $G_0$ and $G_1$, among which the association between the PCF and the protected variable is weak. (C) The initial matches are then refined by considering the between-individual similarities in their baseline characteristics. This step produces the 1-1 counterparts between the subgroups identified by PSM.}} 
    \label{fig:counterpart}
\end{figure}

CFair requires a similarity measurement method for finding counterparts that should handle both confounding issues and systematic differences (see discussion in Section~\ref{sec:factors-influence-dp-gap}). We implemented a two-step approach to achieve this aim, which is illustrated in Fig~\ref{fig:counterpart}. The first step addresses the confounding problem using Propensity Score Matching (PSM). The second step learns a similarity function to ensure that counterparts have similar baseline characteristics.


\subsubsection{Propensity Score Matching}
\label{sec:PSM}
In our case, we utilized PSM to account for the confounding issues between protected and non-protected variables (i.e., some non-protected variables can predict protected variables). We trained an ML model as the propensity score function $PS(\cdot)$ that uses non-protected variables to predict the protected variable under consideration (i.e., the protected variable here is equivalent to the treatment variable in conventional observation studies). The choice of ML model depends on various factors, such as the complexity of confounding effects, availability of data, and imbalanced data issues. We recommend starting with simple models (e.g., logistic regression, decision trees, support vector machine, etc.) and trying more powerful ones (e.g., ensemble models) if more complex relationships between protected and non-protected variables are observed. For individuals whose propensity scores are very close, their protected information cannot be algorithmically distinguished by the model $PS(\cdot)$. Hence, propensity scores can be used to establish initial matches between individuals from different groups (illustrated in Fig~\ref{fig:counterpart}.B) while mitigating confounding bias. More explanations about PSM are provided in Appendix~\ref{sec:PSM-revisted}.

\subsubsection{Identifying 1-1 Counterparts}
\label{subsec:1-1-counterparts}
Since propensity scores are scalars, individuals of similar propensity scores may be diverse in their baseline characteristics represented by non-protected variables. To address this, we introduce an additional similarity measurement after PSM to identify the 1-1 counterparts, and learn the Mahalanobis distance  \cite{de2000mahalanobis} from data to measure similarity between individuals in terms of their baseline characteristics:
\begin{equation}
    s(x, x') = (x - x')^T \boldsymbol{W} (x - x')
\label{equ:dissimilarity-function}
\end{equation}
where $x$ and $x'$ are vectors representing the baseline characteristics of two individuals to be compared, and $\boldsymbol{W}$ is learned from data as explained below. The Mahalanobis distance is well-suited for multivariate data due to its capacity of accommodating correlations and variations in different dimensions or features \cite{wu1997measure,srinivasaraghavan2006application,leys2018detecting}. Users may develop other metric learning approaches tailored to their specific applications, such as those presented in \cite{kaya2019deep, ruoss2020learning, mukherjee2020two, ilvento2020metric, zhao2020immigrate}. We set the learning objective to minimize the total cost of pairwise matching between individuals in the subgroups identified by PSM, the mathematical formulation is shown in the following. Given $x_{n}^{0} \in G_{0}$, we denote $G_{1}^{\prime}(x_{n}^{0} ) \subset G_{1}$ as the set of $\delta$-elements identified by PSM to match with $x_n^0$. We define the cost of matching $x_{n}^{0}$ with $x_{m}^{1} \in G_{1}^{\prime}(x_{n}^{0})$ as follows, which penalizes matching two individuals with distinct baseline characteristics.
\begin{equation}
   c(x_{n}^{0}, x_{m}^{1}) =\alpha_{mn} \ s(x_{n}^{0}, x_{m}^{1})
\end{equation}
The coefficient $\alpha_{mn}$ indicates the probability of $x_{m}^{1}$ being the closest match of $x_{n}^{0}$ and satisfies $\sum_{x_{m}^{1} \in G_{1}^{\prime}(x_{n}^{0})} \alpha_{mn} = 1$, which is also called matching probabilities.  We design $\alpha_{mn}$ as:

\begin{equation}
    \alpha_{mn} = \frac{\exp{ [-s(x_{n}^{0}, x_{m}^{1})}] }{\sum_{x_{k}^{1} \in G_{1}^{\prime}(x_{n}^{0})} \exp{[-s(x_{n}^{0}, x_{k}^{1})] + \epsilon_0}
    }
\end{equation}
where $\epsilon_0$ is a small value (e.g., $10^{-6}$) added to prevent divided by 0. The learning goal is to find $\boldsymbol{W}$ in $s(x, x')$ that minimizes the total cost of pair-wise matching:
\begin{equation}
\begin{aligned}
   \quad C_\text{total} &= \sum_{x_{n}^{0} \in G_{0}} \sum_{x_{m}^{1} \in G_{1}^{\prime}(x_{n}^{0})} c(x_{n}^{0}, x_{m}^{1})  = \sum_{x_{n}^{0} \in G_{0}} \sum_{x_{m}^{1} \in G_{1}^{\prime}(x_{n}^{0})} \alpha_{mn} \ s(x_{n}^{0}, x_{m}^{1})
\end{aligned}
\label{eq:cost-function}
\end{equation}

This is a nonlinear optimization problem. Gradient descent was used to find a suboptimal solution. Once $\boldsymbol{W}$ is decided, we can calculate  ${\alpha_{nm}}$ and then find counterparts in a greedy way.

\section{Experiments}
\label{sec:experiments-all}

We present case studies \footnote{The code is available on \url{https://github.com/zhengyjo/CFair}.} using four datasets widely used in algorithmic fairness studies: the ventilation treatment of sepsis patients in the MIMIC dataset \cite{mimic2022}, which contains critical care information from intensive care units; criminal justice research in the COMPAS dataset \cite{compas2016}; bank credit assessment in the German Credit Dataset \cite{germancredit1994} and income prediction and socioeconomic analysis in the Adult dataset \cite{adult1996}. Appendix~\ref{sec:appendix-data-overview} provides detailed descriptions of the datasets with preprocessing procedures.

\subsection{Significant Systematic Differences Revealed by Propensity Score}
\begin{figure}[t]
    \centering
    \includegraphics[width=0.9\textwidth]{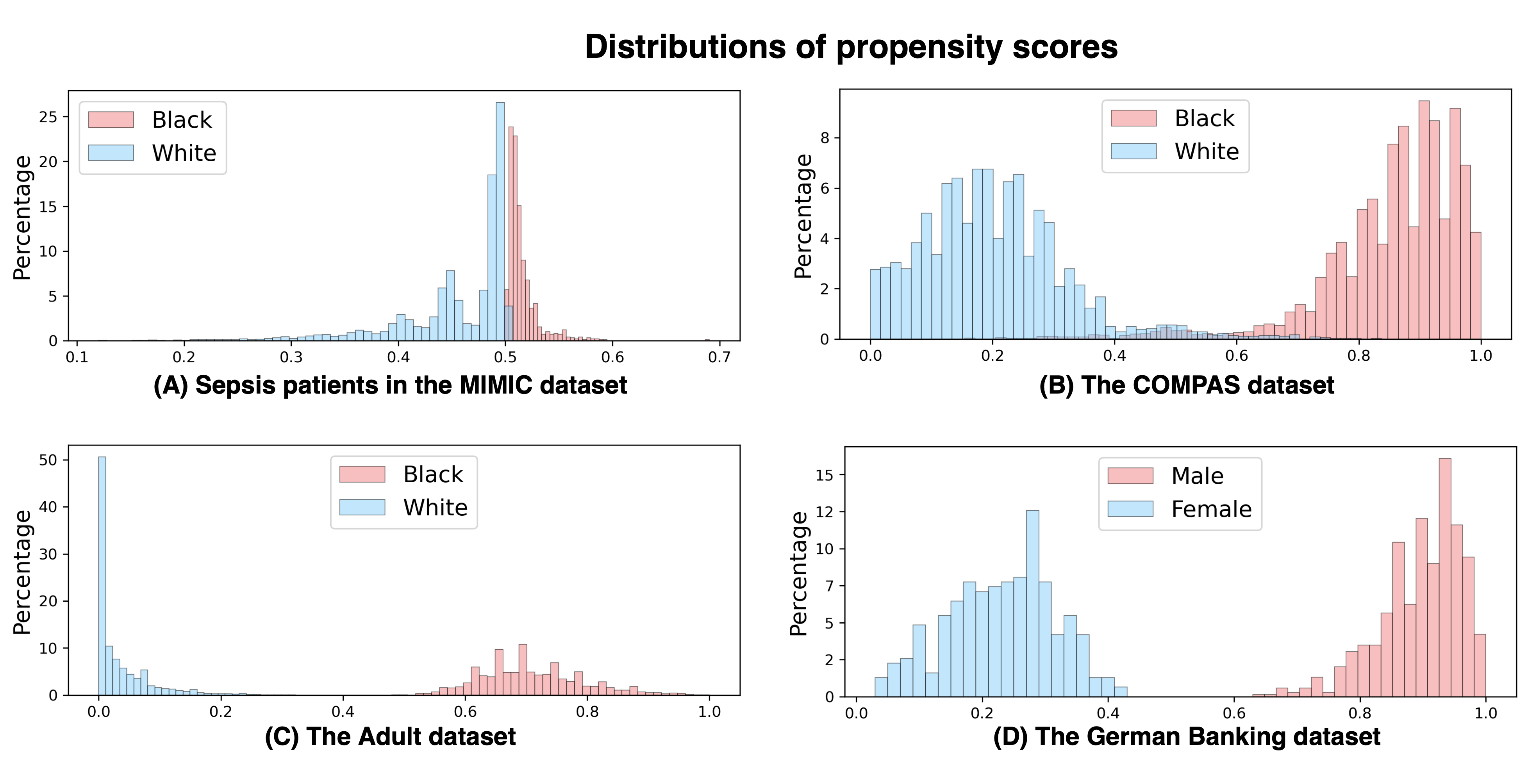}
    \caption{\small{Comparing the propensity score distributions. (A) Black vs White in the case of sepsis patients in the MIMIC dataset, (B) Black vs White in the COMPAS dataset, (C) Male vs Female in the German Banking dataset, and (D) Black vs White in the Adult dataset. Systematic differences are observed in all datasets. Especially, the propensity score distributions of two groups in both (C) and (D) have no overlap at all and their concentrations are well-separated .}}
    \label{fig:ps-bw}
\end{figure}

We tested several machine learning models for calculating propensity scores (details in Appendix~\ref{appendix:build PS models}), and chose AdaBoost \cite{drucker1997improving} (using trees as the base learner) as the propensity score model for the Black vs White case in the MIMIC experiment and random forest for the Black vs White cases in the COMPAS, German Banking, and Adult experiments.  There are far more severe systematic differences in the German Banking and Adult datasets. The propensity score distributions of different race groups are clearly separate (no overlap, distinctive concentrations) (Fig~\ref{fig:ps-bw}(c) and (d)). This reveals the presence of significant systematic differences in these two datasets, pointing to underlying confounding problems that cannot be resolved. Additional supporting evidence can be found in Appendix \ref{sec:appendix-adult-gb}. Hence, we did not include them in the subsequent CFair analysis. For MIMIC and COMPAS, the propensity score distributions of different groups exhibit certain degrees of overlap, implying that CFair can be applied to identify the Black-White counterparts for fairness evaluation. 

\subsection{Fairness Evaluation via CFair}
Using the MIMIC and COMPAS datasets, we demonstrate how to apply CFair to evaluate algorithmic fairness. Note that training a fairer model is not the focus of this study. Random forest was selected as the prediction model  as it achieved the best performance in predicting ventilation status for the sepsis patients in the MIMIC experiment (details in Appendix~\ref{appendix:vent-training})  and predicting recidivism in the COMPAS experiment (details in Appendix~\ref{appendix:recid-training}).

\subsubsection{Mitigation of Systematic Differences}
Table~\ref{tab:t-tests-results} shows that original groups defined in the traditional ways exhibit significant systematic feature-wise differences, which are evidenced by the normalized absolute mean differences of features between groups and the corresponding $t$-test $p$-values (the smaller the more significant). In contrast, the 1-1 counterparts identified by the proposed counterparts selection method are much more similar to each other, evidenced by much smaller differences in features between groups (both the smaller normalized absolute mean differences and the much larger $t$-test $p$-values), which indicates its ability to mitigate the problem of systematic differences. Further examinations of the implementation of counterparts selection, including the two-step matching and the criteria for selecting counterparts, are elaborated in Appendix~\ref{appendix:counterpart-selection-study} and ~\ref{sec:appendix-ablation-study}, respectively.
\begin{table}[t]
\centering
\caption{\small{The t-test p-values comparing feature-wise average between Black and White in MIMIC and COMPAS. Significant systematic differences are observed in the original datasets but are mitigated within the identified counterparts. A smaller $p$-value indicates the difference between the means of the corresponding feature in two racial groups is statistically more significant. Bold formatting highlights P-values deemed significant at the 0.05 significance level. }}
\label{tab:t-tests-results}
\tiny
\setlength{\tabcolsep}{3pt}
\renewcommand{\arraystretch}{1.2}
\scalebox{0.97}{
\begin{tabular}{c|c|c|c|c|c|c|c|c|c|c|c|c|c|c|c|c}
\toprule
\textbf{MIMIC} & Gender & RRT & GCS & Sofa 24h & HR & SBP & DBP & MBP & RR & Temperature & Spo2 & Glucose & Age & CCI & APSiii & BMI \\ \hline
\textbf{Original} & \textbf{<0.001} & \textbf{<0.001} & 0.073 & \textbf{0.015} & \textbf{0.004} & 0.976 & 0.317 & \textbf{<0.001} & \textbf{0.017} & 0.216 & \textbf{<0.001} & 0.554 & \textbf{<0.001} & \textbf{<0.001} & \textbf{<0.001} & 0.462 \\ \hline
\textbf{Counterpart} & 1.000 & 1.000 & 0.922 & 0.599 & 0.720 & 0.064 & 0.666 & 0.800 & 0.975 & 0.741 & 0.992 & 0.542 & 0.361 & 0.786 & 0.759 & 0.734 \\
\bottomrule
\end{tabular}}

\vspace{1em}

\scalebox{1.0}{ 
\setlength{\tabcolsep}{9.5pt}
\begin{tabular}{c|c|c|c|c|c|c|c}
\toprule
\textbf{COMPAS} & Days in Jail & Age & Sex & Decile Score & Priors Count & Days from Compas & V Decile Score \\ 
\hline
\textbf{Original} & \textbf{<0.001} & \textbf{<0.001} & \textbf{<0.001} & \textbf{<0.001} & \textbf{<0.001} & \textbf{<0.001} & \textbf{<0.001} \\ 
\hline
\textbf{Counterparts} & 0.105 & 0.938 & 1.000 & 0.803 & 0.102 & 0.063 & 0.835 \\ 
\bottomrule
\end{tabular}
}
\end{table}
\subsubsection{Compare DP and CDP Gaps}
\label{sec:dp-gap-experiment}
Table~\ref{tab:Dp-gap} summarizes algorithmic fairness analysis using DP gap and CDP gap on ventilation prediction in the MIMIC experiment and recidivism prediction in the COMPAS experiment. The DP gap values were calculated based on the absolute mean difference of the prediction probability between $G_0$ and $G_1$, as indicated in Equ~\ref{equ:dp-gap-estimation}. The CDP gap values were calculated similarly but using the identified counterparts. In both the MIMIC and COMPAS experiments, CFair is able to reveal that the models are statistically significantly biased on the counterparts ($p$-value < 0.001), more severe than what is shown by DP gap. This observation matches the phenomena discussed in \cite{kearns2018preventing} that group fairness may be at the cost of fairness over certain sub-populations. 
 
\begin{table}[t]
\caption{Compare DP and CDP on the ventilation prediction task in the MIMIC experiment and the recidivism prediction task in the COMPAS experiment (5-fold cross-validation)} 
  \label{tab:Dp-gap}
  \small
  \centering
  \begin{tabular}{lccc}
    \toprule
    Dataset & DP gap & CDP gap ($p$-value) & DP Gap - U$^{*}$ \\
    \midrule
    \textbf{MIMIC} & 0.035$\pm$0.007 & 0.058$\pm$0.038 ($<0.001$ ) & 0.048$\pm$0.008 \\
    \textbf{COMPAS} &  0.275$\pm$0.025 & 0.442$\pm$0.106 ($<0.001$ ) & 0.218$\pm$0.028\\
    \bottomrule
    \scriptsize{$^{*}$Unmatched population}
  \end{tabular}
  
\end{table}

\begin{table}[t]
\caption{Fairness analysis using Equal Opportunity and Sufficiency on ventilation prediction task from MIMIC and the recidivism prediction task from COMPAS (5-fold cross-validation, random forest as the prediction model). Two racial groups (Black and White) are considered.}
\label{tab:mimic-compas-merge}
  \scriptsize
  \centering
  \begin{tabular}{l|cc|cc|cc}
    \toprule
    \textbf{MIMIC} &  \multicolumn{2}{c}{Counterparts} & \multicolumn{2}{c}{Unmatched population} & \multicolumn{2}{c}{Total population}   
    \\
   &  \scriptsize{Black} & \scriptsize{White} &  \scriptsize{Black} & \scriptsize{White} &  \scriptsize{Black} & \scriptsize{White} \\
    \midrule
\scriptsize{\textbf{Accuracy}} & \scriptsize{0.737$\pm$0.053} & \scriptsize{0.772$\pm$0.044} & \scriptsize{0.685$\pm$0.028} & \scriptsize{0.727$\pm$0.006} & \scriptsize{0.700$\pm$0.016} & \scriptsize{0.728$\pm$0.006} \\
\midrule
\scriptsize{\textbf{Equal Opportunity ($\Delta\text{TPR}$)}} & \multicolumn{2}{c|}{0.266$\pm$0.204 (\textit{p}-value $<0.001$)} & \multicolumn{2}{c|}{0.104$\pm$0.059} & \multicolumn{2}{c}{0.102$\pm$0.075}  \\
\midrule
\scriptsize{\textbf{Sufficiency ($\Delta \text{PPV}$)}} & \multicolumn{2}{c|}{0.477$\pm$0.275 (\textit{p}-value $<0.001$)} & \multicolumn{2}{c|}{0.185$\pm$0.085} & \multicolumn{2}{c}{0.158$\pm$0.085}  \\
    \bottomrule
  \end{tabular}

\vspace{1em}
  
  \begin{tabular}{l|cc|cc|cc}
    \toprule
 \textbf{COMPAS}   &  \multicolumn{2}{c}{Counterparts} & \multicolumn{2}{c}{Unmatched population} & \multicolumn{2}{c}{Total population}   
    \\
   &  \scriptsize{Black} & \scriptsize{White} &  \scriptsize{Black} & \scriptsize{White} &  \scriptsize{Black} & \scriptsize{White} \\
    \midrule
\scriptsize{\textbf{Accuracy}} & \scriptsize{0.670$\pm$0.079} & \scriptsize{0.876$\pm$0.039} & \scriptsize{0.624$\pm$0.011} & \scriptsize{0.672$\pm$0.013} & \scriptsize{0.629$\pm$0.017} & \scriptsize{0.701$\pm$0.009} \\
\midrule
\scriptsize{\textbf{Equal Opportunity ($\Delta \text{TPR}$)}}  & \multicolumn{2}{c|}{0.435$\pm$0.119 (\textit{p}-value $<0.001$)} & \multicolumn{2}{c|}{0.191$\pm$0.026} & \multicolumn{2}{c}{0.162$\pm$0.024}  \\
\midrule

\scriptsize{\textbf{Sufficiency ($\Delta \text{PPV}$)}}  & \multicolumn{2}{c|}{0.366$\pm$0.176 (\textit{p}-value $<0.001$)} & \multicolumn{2}{c|}{0.085$\pm$0.053} & \multicolumn{2}{c}{0.086$\pm$0.043}  \\

    \bottomrule
  \end{tabular}
  \label{tab:compas}
\end{table}

\subsubsection{Adaption of CFair to More Group Fairness Indexes}
\label{sec:CFair_other_indexes}
CFair can be applied to other group fairness indexes (e.g., Equal Opportunity \cite{barocas2023fairness,lee2021fair}, Sufficiency \cite{castelnovo2022clarification}), etc., which examine fairness from different viewpoints. Table~\ref{tab:mimic-compas-merge} demonstrate this generalizability using the ventilation prediction task from MIMIC  and the recidivism prediction task from COMPAS. Equal Opportunity is measured by $\Delta\text{TPR}$ (the group-wise difference of the true positive rate), and Sufficiency is measured by $\Delta\text{PPV}$ (the group-wise difference of the positive predictive value). It is shown that CFair (the "Counterparts" column) is able to reveal deeper algorithmic unfairness potentially ignored by both Equal Opportunity and Sufficiency in their original forms (the "Total population" column). For example, in the MIMIC experiment, the bias quantified by the vanilla Equal Opportunity (0.102$\pm$0.075) is much milder than then the one quantified by CFair (0.266$\pm$0.204). In addition, CFair is able to assess the statistical significance ($p$-value < 0.001) of the bias quantity. Similar results are obtained if Sufficiency is used. More supporting results are provided in Appendix \ref{appendix:additional-res-mimic} for the MIMIC experiment and in Appendix \ref{appendix:extra-disscuss-compas} for the COMPAS experiment. These observations confirm that strong signals of bias against similar individuals from different groups can be dramatically diluted by the signals produced by other individuals in the total population.

\section{Discussion and Conclusion}
Fairness assessment in machine learning applications is critical across sectors to ensure their equitable and effective contributions to decision-making.  It is essential to rigorously and continuously monitor these ML models to prevent discrimination against specific individuals or groups. This vigilance and commitment to fairness extend to other important sectors, including finance for bank loan approvals, the criminal justice system for incarceration decisions, and employment for hiring practices \cite{zhang2018fairness}. Conducting fairness analysis is vital for enhancing  the trustworthiness of ML models in diverse real-world settings, and ultimately benefit society at large. However, there is a gap where systematic differences are overlooked in the above discussed cases by existing fairness metrics. In this work, we provide both theoretical and empirical evidence that systematic differences in data can dramatically affect the results of fairness evaluations. To mitigate the negative impacts of systematic differences, we introduce CFair, which evaluates algorithmic fairness on 1-1 counterparts. The 1-1 counterparts are individuals with similar baseline characteristics regardless of protected variables. We have developed a method that uses propensity score matching and metric learning for identifying 1-1 counterparts. Experimental results show that this method is effective in finding similar individuals. 

Moreover, we have shown the importance of assessing the statistical significance of a quantified disparity magnitude, regardless of the chosen fairness metrics. Without a meticulous statistical assessment, it is challenging to discern whether a disparity value stems from randomness. CFair offers a novel means to perform this evaluation, which is not available in existing fairness evaluation methods. This is currently implemented by using the paired $t$-tests on the identified counterparts. It is shown in Section~\ref{sec:dp-gap-experiment} and Section~\ref{sec:CFair_other_indexes} that a model can be significantly biased even though it only produces a small quantity of disparity. It should be noted that CFair is not against existing fairness metrics. Instead, as demonstrated in the experimental results, CFair offers a novel way to use existing fairness metrics (e.g., DP gap, Equal Opportunity, and Sufficiency) to reveal fairness issues that may be ignored by conventional fairness analysis. This is reverberated in an experiment using synthetic data (see Appendix \ref{appendix:synthetic_experiment}), in which the ground-truth (both class labels and counterparts) is known. The findings also indicate that, to enhance both model performance and algorithmic fairness, it could be advantageous to identify regions exhibiting systematic differences and construct a model tailored to each specific region. Although it may not always be feasible to create a custom model for every single region in practical scenarios (e.g., due to various data issues), it remains essential to identify and address regions where bias is evident in the model. Finally, it is worth highlighting that it is straightforward to extend CFair analysis to use other fairness metrics beyond those showcased in this study to encompass a wider array of metrics. We leave the discussion of future work in Appendix~\ref{appendix:future}.


\section*{Acknowledgment}
This work is funded by NIH/NLM 1r01lm014239.

{\small
\bibliography{neurips_2023}}
\bibliographystyle{nips2}

\newpage
\appendix
\counterwithin{figure}{section}
\counterwithin{equation}{section}
\counterwithin{table}{section}

\section{Related Works}
\label{sec:related-works}
\paragraph{Group Fairness in Machine Learning.}
Many group fairness metrics have been proposed to quantify the fairness of algorithmic decisions and help train fairer ML models. The choice of fairness evaluation metrics will depend on the specific usage and the desired level of fairness. Popular metrics include equal odds \cite{hardt2016equality}, Equal Opportunity \cite{zafar2017fairness}, treatment equality \cite{berk2021fairness}, equal allocation like demographic parity \cite{zhao2019inherent}, and so on.
Group fairness has been widely employed in machine learning models. Related studies on fair representation learning use autoencoder \cite{madras2018learning}, adversarial training \cite{zhao2019inherent, zhao2019conditional}, optimal transport \cite{gordaliza2019obtaining, jiang2020wasserstein}, or fair kernel methods \cite{donini2018empirical} to remove any information relevant to the protected variables while preserving as much information as possible for downstream tasks. However, most group fairness measurements use groups defined by protected variables and under-recognize the effects of systematic between-group differences in the baseline characteristics related to the tasks under consideration.

\paragraph{Causality-based Fairness.} 
Causality-based criteria try to employ domain or expert knowledge to come up with a causal structure of the problem. Mitchell et al.~\cite{mitchell2021algorithmic} reviewed various choices and assumptions of fairness in decision-making, including a clear investigation of causal definitions. For instance, counterfactual fairness \cite{kusner2017counterfactual} assumes that an individual's prediction outcome remains constant after changing the values of protected variables, which is subject to strong assumptions about the data and the underlying mechanism generating them. Kilbertus et al.~\cite{kilbertus2017avoiding} theoretically showed how to avoid unresolved discrimination and proxy discrimination by making interventions on the causal graph. Zhang et al.~\cite{zhang2018fairness} extended the causal concepts introduced in \cite{blank2004measuring} and partitioned the total disparity into disparities from each type of path (direct, indirect, back-door), and derived the causal explanation formulas accordingly. Causality-based fairness methods are preferable in principle but challenging to implement in applications without clear causal structures~\cite{castelnovo2022clarification, mitchell2021algorithmic}.  

\paragraph{Fairness in Electronic Health Records Data.} The medical data usually have representation bias and aggregation bias problems due to complex historical structural inequities and social determinants in healthcare. Many ML approaches have been found to show discrimination towards certain demographic groups when applied to analyze EHR data \cite{chen2020exploring}. For example, it was found that prediction models trained with the MIMIC dataset unequally relied on racial attributes across subgroups \cite{meng2022interpretability}. Similar differences are also observed among groups with different genders, marital status, or insurance types \cite{buolamwini2018gender, salles2019estimating, stepanikova2008effects}. Another example \cite{pfohl2019creating} showed that fairness issues occurred in the use of pooled cohort equations (PCE), which guides physicians in deciding whether to prescribe cholesterol-lowering therapies to prevent ASCVD. Its observational study indicates that PCE tends to overestimate risk, putting different groups at risk of being under- or over-treated. The above observations clearly indicate the importance of fairness analysis in developing ML models and techniques for healthcare applications.

\section{Discussion on DP Gap Estimation}
\label{appendix:biased-sampling}
In this section, we study how data sampling affects the estimation of DP gap. Since this theoretical analysis does not account for the error of prediction, we assume $\hat{Y}=Y$ and use the notation $Y$ in the remaining text.

We assume the samples of two groups come from two independent underlying marginal distributions: $\phi_{0}(Y)$ and $\phi_{1}(Y)$. We suppose there are $N_0$ iid random samples $Y_{1}^{(0)}, Y_{2}^{(0)}, ..., Y_{N_0}^{(0)} \sim \phi_{0}(Y)$ and $N_1$ iid samples $Y_{1}^{(1)}, Y_{2}^{(1)}, ..., Y_{N_1}^{(1)} \sim \phi_{1}(Y)$. We use notation of $\bar{Y}_{N_0}=\sum_{i=1}^{N_0}Y_{i}^{(0)}/N_{0}$ and $\bar{Y}_{N_1}=\sum_{j=1}^{N_1}Y_{j}^{(1)}/N_{1}$ to represent the sample average of each group, respectively. Here we have $\mathbb{E}[\bar{Y}_{N_0}]=\mathbb{E}[\phi_{0}(Y)]=\nu_0$ and $\mathbb{E}[\bar{Y}_{N_1}]=\mathbb{E}[\phi_{1}(Y)]=\nu_1$. Without loss of generality, we also assume two marginal distributions have the same variance, that is, $\text{Var}[\phi_{0}(Y)]=\text{Var}[\phi_{1}(Y)]=\sigma^{2}$. 
In what follows we aim to derive the probability density function (PDF) of DP gap: $\widehat{\Delta_{\text{DP}}}=|\bar{Y}_{N_0}-\bar{Y}_{N_1}|$.

According to Central Limit Theorem, we have 
\begin{equation}
    \bar{Y}_{N_0} \sim \mathcal{N}(\nu_0, \frac{\sigma^2}{N_0}) \quad \text{as} \ N_0 \rightarrow +\infty,\quad \text{and}\quad \bar{Y}_{N_1} \sim \mathcal{N}(\nu_1, \frac{\sigma^2}{N_1}) \quad \text{as} \ N_1 \rightarrow +\infty.
\end{equation}

The following derivation is based on the approximation of Central Limit Theorem.
Since $\bar{Y}_{N_0}$ and $\bar{Y}_{N_1}$ are independent random variables that are normally distributed, we have
\begin{equation}
\bar{Y}_{N_0}-\bar{Y}_{N_1} \sim \mathcal{N}(\Delta \nu, \sigma_{1}^{2})\end{equation}
where $\Delta \nu=\nu_0 - \nu_1$ and $\sigma_{1}^{2}=\frac{\sigma^2}{N_0}+\frac{\sigma^2}{N_1}$. Note that $\widehat{\Delta_{\text{DP}}}$ is the folded normal distribution of $|\bar{Y}_{N_0}-\bar{Y}_{N_1}|$, its PDF is given by
\begin{equation}
    f_{\widehat{\Delta_{\text{DP}}}}(x)=\frac{1}{\sqrt{2\pi\sigma_{1}^{2}}}e^{-\frac{(x-\Delta \nu)^{2}}{2\sigma_{1}^{2}}} + \frac{1}{\sqrt{2\pi\sigma_{1}^{2}}}e^{-\frac{(x+\Delta \nu)^{2}}{2\sigma_{1}^{2}}}, \ \text{for} \ x \ge 0, 
\end{equation}
and 0 everywhere else. According to \cite{leone1961folded, tsagris2014folded}, the expectation and variance are expressed as 
\begin{equation}
    \mathbb{E}[\widehat{\Delta_{\text{DP}}}]=\sqrt{\frac{2}{\pi}}\sigma_1 e^{-\frac{(\Delta \nu)^{2}}{2\sigma_{1}^{2}}}+\Delta \nu[1-2\Phi(-\frac{\Delta \nu}{\sigma_{1}})], \ \text{Var}[\widehat{\Delta_{\text{DP}}}]=(\Delta \nu)^{2}+\sigma_{1}^{2}-\mathbb{E}[\widehat{\Delta_{\text{DP}}}]^{2}
\label{equ:dp-gap-expectation}
\end{equation}
where $\Phi$ is the normal cumulative distribution function.
\paragraph{Effects of Unbalanced Data Sampling}
In this scenario, we assume samples from two groups are fully representative but with unbalanced sampling $N_{0} \ll N_{1}$, in other words, $\frac{N_0}{N_1}=o(1)$. Without loss of generality, we assume that $\Delta \nu = 0$ and thus $\Delta_{\text{DP}}=0$. \footnote{If $\Delta \nu \neq 0$, we could introduce the modified samples $\Tilde{Y}_{i}^{(1)}=Y_{i}^{(1)}-\Delta \nu$ and the modified average $\Bar{\Tilde{Y}}_{N_1}=\Tilde{Y}_{N_1} -\Delta \nu$. Estimating DP gap error based on $\Tilde{Y}_{i}^{(0)}$ and $\Tilde{Y}_{i}^{(1)}$ reduces to the case where $\Delta \nu = 0$.} Then we have $\mathbb{E}[\widehat{\Delta_{\text{DP}}}]=\sqrt{\frac{2}{\pi}}\sigma_1, \text{Var}[\widehat{\Delta_{\text{DP}}}]=(1-\frac{2}{\pi})\sigma_{1}^2$. Recall the alternative formulation of $\sigma_{1}^{2}=\frac{\sigma^{2}}{N_{0}}(1+\frac{N_0}{N_1})$ and  $\frac{N_0}{N_1}=o(1)$, we have $\sigma_{1}^{2} \approx \frac{\sigma^{2}}{N_0}$. $\mathbb{E}[\widehat{\Delta_{\text{DP}}}]$ and $\text{Var}[\widehat{\Delta_{\text{DP}}}]$ are both inversely proportional to $N_0$, indicating that the unbalanced data sampling would undermine the precision and stability in DP gap estimation.

\paragraph{Effects of Biased Sampling.} The deviation of $\mathbb{E}[\widehat{\Delta_{\text{DP}}}]$ is positive correlated to $|\Delta \nu|$, which is derived from the first derivative of $\mathbb{E}[\widehat{\Delta_{\text{DP}}}]$. 
\begin{equation}
\begin{aligned}
    \frac{d\mathbb{E}[\widehat{\Delta_{\text{DP}}}]}{d(\Delta \nu)} & =-\sqrt{\frac{2}{\pi}}\frac{\Delta \nu}{\sigma_1}e^{-\frac{(\Delta \nu)^2}{2\sigma_{1}^{2}}} + 1 -2[\Phi(-\frac{\Delta \nu}{\sigma_1})-\frac{\Delta \nu}{\sqrt{2\pi}\sigma_1}e^{-\frac{(\Delta \nu)^2}{2\sigma_{1}^{2}}}] \\
    & = 1-2\Phi(-\frac{\Delta \nu}{\sigma_{1}^{2}})
\end{aligned}
\label{equ:dp-gap-derivative}
\end{equation}
According to Equation~\ref{equ:dp-gap-derivative}, $\frac{d\mathbb{E}[\widehat{\Delta_{\text{DP}}}]}{d(\Delta \nu)} < 0$ if $\Delta \nu <0$, and $\frac{d\mathbb{E}[\widehat{\Delta_{\text{DP}}}]}{d(\Delta \nu)} > 0$ if $\Delta \nu >0$. $\mathbb{E}[\widehat{\Delta_{\text{DP}}}]$ increases as $|\Delta \nu|$ increases and it achieves minimum of $\sqrt{\frac{2}{\pi}}\sigma_1$ if and only if $\Delta \nu = 0$. Biased sampling would result in a nonzero $\Delta_v$, which could either overestimate or underestimate the DP gap.
\section{Revisiting Propensity Score Matching}
\label{sec:PSM-revisted}
Primarily used in randomized control trials, propensity score is the probability of treatment assignment conditional on observed baseline characteristics \cite{austin2011introduction}. In observational studies, randomly assigning participants to treatment groups can lead to confounding issues and biased estimates of treatment effects. Propensity score matching (PSM) effectively addresses the above issues by balancing covariate distributions between treatment groups, enabling a more accurate estimation of treatment effects in non-randomized settings. PSM is a valuable tool commonly used in various domains, such as, social sciences, economics, healthcare and medicine, public policy and program evaluation, education, and so on. In this work, we extends it to the case of race assignment and utilize PSM to mitigate confounding issues when identifying comparable individuals between two groups. In what follows we explain a few ways for deciding propensity score using machine learning models.

One common way to determine propensity scores is by fitting a logistic regression using selected covariates to predict the values of a sample's protected variables. Assuming the protected variable is binary (taking 0 or 1), the probability of it equal to 1 is $p = (1 + e^{-(\beta_0+\beta_{1}x_{1}+...+\beta_{m}x_{m})})^{-1}$, where $\beta_0$ is the intercept coefficient and $\beta_1, ..., \beta_m$ are the regression coefficients for the $m$ selected covariates $x_1, ..., x_m$. The propensity score will be: $s = \log(\frac{p}{1-p})$. Other machine learning models, such as decision trees \cite{loh2011classification} and its ensemble models \cite{drucker1997improving},  support vector machine \cite{hearst1998support} and neural network \cite{popescu2009multilayer}, can also be used to determine propensity scores. These models can capture complex relationships between features, potentially leading to more accurate propensity score estimation.

Two individuals from different groups will be matched if their propensity scores (i.e., $s_0, s_1$) are close enough, that is, $\Delta_s := |s_0-s_1|<\delta$. To determine the threshold $\delta$ of PSM, in this work we gathered the empirical distributions of $\Delta_s$ over all possible pairs and set the 90-th percentile as the threshold.

\section{Broader Impacts, Limitations and Extensions} 
\label{appendix:future}
Integration of machine learning with clinical decision support tools, like diagnostic support, has the potential to improve clinical decisions by providing targeted and timely information to healthcare providers \cite{gianfrancesco2018potential}. Practically this means deployed algorithms must be monitored to ensure their safety and to avoid discrimination against specific individuals or groups. This concern extends beyond healthcare scenarios to other critical areas such as bank loans, criminal incarceration, and employment decisions \cite{zhang2018fairness}. Moving forward to fairness study would improve the trustworthiness of machine learning models in real-world applications and benefit society as a whole. In this work, we first theoretically analyze the impact of systematic differences on fairness evaluation using demographic parity as a paradigmatic example, and then provide empirical evidence. Our finding offers insights towards implementing trustworthy matching learning systems, that is, encouraging researchers to consider systematic differences and confounding effects in evaluating algorithmic fairness. 

CFair depends on the possibility of identifying a sufficient number of counterparts, which may be difficult in some applications. In such scenarios, one may make inappropriate conclusions using CFair, and should investigate the underlying reasons and potential ways for making improvements. If the problem is due to overwhelming systematic differences in data, one may seek to improve the data collection process. We have shown in this work how to reveal systematic differences in data by comparing between-group feature-wise divergence. If the problem is due to the utilization of an inappropriate method for measuring similarity between individuals or matching individuals, endeavors can be devoted to improving the similarity measurement method and/or the matching method. In practice, it is possible to match outliers as counterparts although the likelihood is minimal when an appropriate similarity measurement is deployed. We expect the number of outlier counterparts to be significantly smaller than that of normal counterparts, and the effect of outlier counterparts on fairness analysis should be very limited. 

In our current implementation, we constrain CFair to use 1-1 counterparts, which sometimes is necessary to mitigate systematic differences (see Appendix~\ref{appendix:counterpart-selection-study}). However, this is case-dependent. In other applications, one individual from a group may be similar to several individuals from another group (i.e., one individual may have multiple counterparts) without incur systematic differences. It could be a missed opportunity to use only 1-1 counterpart in fairness analysis. Exploring one-to-many matching approaches in CFair analysis can prove to be a promising avenue for future research.

\section{Proofs}
\label{sec:appendix-CFair}
\textbf{Corollary 3.3.} \textit{Given two groups $G_0$ and $G_1$, both $C_{0, \delta}$ and $C_{1, \delta}$ are unique.}
\begin{proof}
    We give proof of the uniqueness of $C_{0, \delta}$ by contradiction. Suppose two distinct groups $C_{0, \delta}^{a}$ and $C_{0, \delta}^{b}$ are both the $\delta$-groups of $G_0$. Then there exists an element $x\in C_{0, \delta}^{b}$ but not in $C_{0, \delta}^{a}$. According to Definition~\ref{def:counterpart-subgroup}, $x$ is a $\delta$-element and thus $x$ also belongs to $C_{0, \delta}^{a}$, which leads to contradiction. The proof of $C_{1, \delta}$ is the same. 
\end{proof}

\section{Additional Experimental Results}
\subsection{Datasets Overview}
\label{sec:appendix-data-overview}
In this work, we considered four datasets of diverse applications in fairness analysis: MIMIC \cite{mimic2022}, Adult \cite{adult1996}, German Banking \cite{germancredit1994} and COMPAS \cite{compas2016}. 

\begin{figure}[t]
    \centering
    \includegraphics[width=1.0\textwidth]{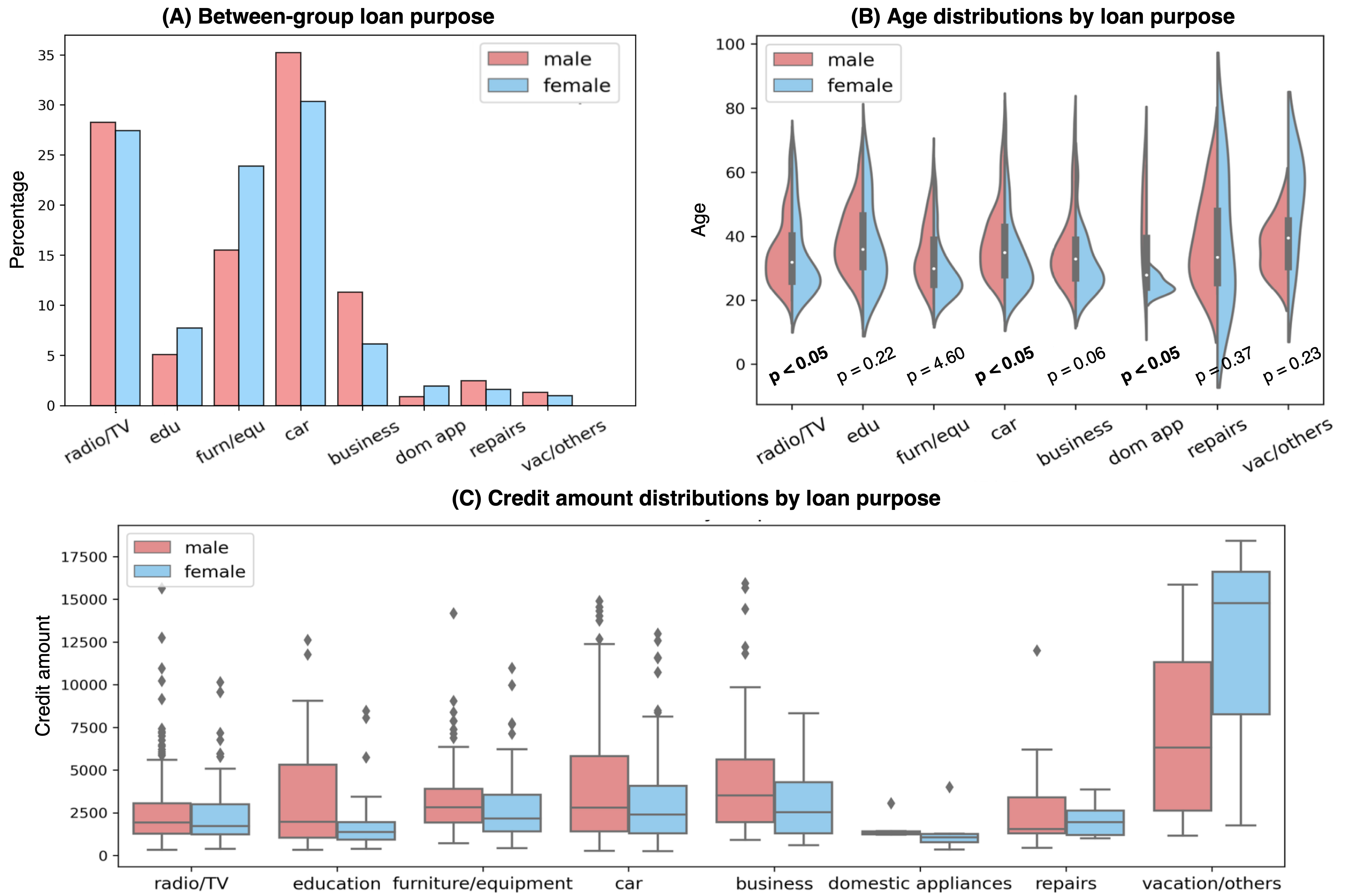}
 \caption{Systematic differences are observed between males and females in the German Banking dataset \cite{germancredit1994}. The target is to predict the risk of a client ("good" vs "bad"). (A) Males and females have different loan purposes. Hence, it is reasonable for banks to treat their loan applications differently, for example, by enforcing different levels of scrutiny and requiring different documentations. (B) The age distributions of male and female applicants are statistically significantly different (the Kolmogorov-Smirnov test \cite{massey1951kolmogorov} $p$-value < 0.05) in several loan categories: radio/TV, car, and domestic appliance. Since age is an important factor in making loan decisions by lenders, it is expected that banks would treat the applications from males and females differently even if they have the same loan purpose. (C) The credit amount distributions also significantly differ between males and females in several borrowing purpose categories. In general, females had a lower average credit compared to males across many load purpose categories.}
    \label{fig:german-feature-difference}
\end{figure}

\begin{itemize}
\item The MIMIC dataset \cite{mimic2022} contains critical care data for patients admitted to intensive care units at the Beth Israel Deaconess Medical Center. In this experiment, we focused on the task of building a model to predict the ventilation status of sepsis patients. Table~\ref{tab:sepsis-feature} lists the features used in this study, which includes demographics, laboratory test results, and treatments. For each patient, we only used the first medical records (i.e., at 0th hour) from each individual's first ICU visit to avoid potential treatment biases. Outliers were removed by checking each of the following features: Age, BMI, Temperature, HR, RR, SBP, DBP, SpO2, MBP, Glucose. Specifically, we removed a patient if at least one of the features fell outside of the [2.5, 97.5] percentile of the total population. When we studied Black vs White, the total population consisted of the Black and White patients. After removing outliers, we had 990 Black patients and 9,254 White patients.

\item The Adult Income dataset \cite{adult1996}, commonly known as the Adult dataset, is a resource for studying income prediction and socioeconomic factors. It contains the demographic features of individuals (e.g., age, education, occupation, marital status, etc.) and whether their income exceeds a certain threshold. Researchers frequently utilize this dataset to explore various machine-learning algorithms and techniques for predicting income levels. Moreover, the Adult dataset has been widely used to to study algorithmic bias. In our experiment, we extracted the Black and the White groups (4,685 Black individuals and 41,762 White individuals). 

\item The German Banking dataset \cite{germancredit1994}, also referred to as the German Credit dataset, is a classic benchmark for developing and evaluating credit risk assessment models. It comprises information about credit applicants and whether they defaulted on their credit obligations. The features include various financial attributes, such as credit amount, duration, and installment rate, as well as personal information. Fig~\ref{fig:german-feature-difference} illustrates the systematic differences between males and females.

\item The Correctional Offender Management Profiling for Alternative Sanctions (COMPAS) dataset \cite{compas2016} is a dataset frequently used in criminal justice research (especially, biases and disparities in criminal justice decision-making). It encompasses data from the criminal justice system and contains demographic information, criminal history, and assessment results. The target is to build recidivism prediction models and evaluate algorithmic fairness. In this work, we focused on the African-American (5,153 individuals) and Caucasian (3,579 individuals) groups, where individuals with missing target labels were excluded.
\end{itemize}

\begin{table}[h]
    \centering
    \begin{tabular}{p{0.13\linewidth} | p{0.80\linewidth}}
    \toprule
 \textbf{Features} &  \textbf{Description}  \\
 \midrule
 Age & The age of a patient upon admission to the ICU. \\
 \midrule
BMI & Body mass index.\\
 \midrule
Gender & The individual's gender.\\
 \midrule
Temperature & The body temperature.\\
 \midrule
HR & Heart rate. \\
 \midrule
RR & Respiratory rate.\\
 \midrule
SBP & Systolic blood pressure.  \\
 \midrule
DBP & Diastolic blood pressure.\\
 \midrule
SpO2 & Oxygen saturation. \\
 \midrule
MBP & Mean blood pressure.\\
 \midrule 
Glucose & Blood glucose level. \\
 \midrule
 CCI & Charlson Comorbidity Index \cite{roffman2016charlson}. It indicates the mortality risk within 1 year of hospitalization.\\
 \midrule
APSiii & The APACHE-III score \cite{darbandsar2016acute}. It estimates a given patient's severity of illness .\\
 \midrule
 RRT & Renal Replacement Therapy \cite{miller2003rrt}.  It's a binary indicator for kidney function replacement. \\
 \midrule
 GCS & Glasgow Coma Scale \cite{sternbach2000glasgow}. It describes the extent of impaired consciousness. \\
 \midrule
SOFA & Sequential Organ Failure Assessment \cite{vincent1998use}. It assesses the performance of a body's organ systems.  \\
 \midrule
\vspace{0.03cm} Ventilation & Ventilation treatment. It is a categorical variable with 3 classes: no ventilation, supplemental oxygen, and invasive ventilation. \\
\bottomrule
\end{tabular}
\caption{Feature codebook of sepsis patients in MIMIC \cite{mimic2022}}
\label{tab:sepsis-feature}
\end{table}

\subsection{Additional Results of the MIMIC Experiment}
\subsubsection{Overlooked Bias Issues Uncovered by CFair.}
\label{appendix:additional-res-mimic}
CFair is able to reveal significant algorithmic biases that may be ignored by the vanilla fairness analyses using Equal Opportunity based on True Positive Rate and Sufficiency based on Positive Predictive Value. The Equal Opportunity value on the total population is $0.102$, which is less than half of that on the counterparts ($0.266$). Similarly, The Sufficiency value on the total population is $0.158$, only about one third of that on the counterparts ($0.477$). Digging into each ventilation category (i.e., the target of the prediction model), we can see the composition details of the overall performance discrepancies. For example, in the "Supplemental Oxygen" category, the model generates the most disparity: the Equal Opportunity value on the total population is less than one third of that on the counterparts, and the Sufficiency value on the total population is only about one third of that on the counterparts. In the "Invasive Ventilation" category, the model generates less, but still noticeable, disparity on the counterparts than on the total population. This shows that focusing on specific subpopulation can greatly affect fairness analysis results. In addition, it can provide directions for improving the ML model or designing a better (or more trustworthy) way to use the current ML model (e.g., constraining the scenarios that it can be applied to).

\begin{table}[h] 

  \caption{Fairness analysis using Equal Opportunity and Sufficiency on ventilation prediction for the sepsis patients in the MIMIC dataset (5-fold cross-validation, random forest as the prediction model). Two groups (Black and White) are considered. $\Delta{\text{TPR}}$ and  $\Delta{\text{PPV}}$ are the absolute group-wise differences in TPR and PPV, respectively. The subscripts indicate classes: No Ventilation (NV); Supplemental Oxygen (SO); and Invasive Ventilation (IV).}
  \scriptsize
    \label{tab:mimic-metric-detail}
  \centering
  \begin{tabular}{l|cc|cc|cc}
    \toprule
    &  \multicolumn{2}{c}{Counterparts} & \multicolumn{2}{c}{Unmatched population} & \multicolumn{2}{c}{Total population}   
    \\
   &  \scriptsize{Black} & \scriptsize{White} &  \scriptsize{Black} & \scriptsize{White} &  \scriptsize{Black} & \scriptsize{White} \\
    \midrule
\textbf{Accuracy} & \scriptsize{0.737$\pm$0.053} & \scriptsize{0.772$\pm$0.044} & \scriptsize{0.685$\pm$0.028} & \scriptsize{0.727$\pm$0.006} & \scriptsize{0.700$\pm$0.016} & \scriptsize{0.728$\pm$0.006} \\
\midrule
\scriptsize{\textbf{Equal Opportunity}} & \multicolumn{2}{c|}{0.266$\pm$0.204} & \multicolumn{2}{c|}{0.104$\pm$0.059} & \multicolumn{2}{c}{0.102$\pm$0.075}  \\
$\Delta{\text{TPR}}_{\text{NV}}$ &  \multicolumn{2}{c|}{0.051$\pm$0.036} & \multicolumn{2}{c|}{0.021$\pm$0.017} & \multicolumn{2}{c}{0.024$\pm$0.010} \\
$\Delta{\text{TPR}}_{\text{SO}}$ &  \multicolumn{2}{c|}{0.227$\pm$0.203} & \multicolumn{2}{c|}{0.069$\pm$0.027} & \multicolumn{2}{c}{0.069$\pm$0.029} \\
$\Delta{\text{TPR}}_{\text{IV}}$ &  \multicolumn{2}{c|}{0.122$\pm$0.142} & \multicolumn{2}{c|}{0.082$\pm$0.073} & \multicolumn{2}{c}{0.082$\pm$0.085} \\
\midrule
\scriptsize{\textbf{Sufficiency}} & \multicolumn{2}{c|}{0.477$\pm$0.275} & \multicolumn{2}{c|}{0.185$\pm$0.085} & \multicolumn{2}{c}{0.158$\pm$0.085}  \\
$\Delta{\text{PPV}}_{\text{NV}}$ &  \multicolumn{2}{c|}{0.047$\pm$0.023} & \multicolumn{2}{c|}{0.052$\pm$0.035} & \multicolumn{2}{c}{0.032$\pm$0.023} \\
$\Delta{\text{PPV}}_{\text{SO}}$ &  \multicolumn{2}{c|}{0.521$\pm$0.291} & \multicolumn{2}{c|}{0.178$\pm$0.093} & \multicolumn{2}{c}{0.180$\pm$0.091} \\
$\Delta{\text{PPV}}_{\text{IV}}$ &  \multicolumn{2}{c|}{0.190$\pm$0.119} & \multicolumn{2}{c|}{0.072$\pm$0.027} & \multicolumn{2}{c}{0.067$\pm$0.031} \\
\midrule
$\text{TPR}_{\text{NV}}$ & \scriptsize{0.909$\pm$0.043} & \scriptsize{0.913$\pm$0.026} & \scriptsize{0.909$\pm$0.022} & \scriptsize{0.900$\pm$0.005} & \scriptsize{0.910$\pm$0.022} & \scriptsize{0.900$\pm$0.006} \\
$\text{TPR}_{\text{SO}}$ & \scriptsize{0.080$\pm$0.160} & \scriptsize{0.307$\pm$0.355} & \scriptsize{0.057$\pm$0.071} & \scriptsize{0.094$\pm$0.011} & \scriptsize{0.063$\pm$0.053} & \scriptsize{0.096$\pm$0.011} \\
$\text{TPR}_{\text{IV}}$ & \scriptsize{0.148$\pm$0.140} & \scriptsize{0.270$\pm$0.157} & \scriptsize{0.187$\pm$0.058} & \scriptsize{0.269$\pm$0.018} & \scriptsize{0.187$\pm$0.076} & \scriptsize{0.269$\pm$0.019} \\
\midrule
$\text{PPV}_{\text{NV}}$ & \scriptsize{0.801$\pm$0.050} & \scriptsize{0.826$\pm$0.069} & \scriptsize{0.736$\pm$0.041} & \scriptsize{0.785$\pm$0.008} & \scriptsize{0.755$\pm$0.022} & \scriptsize{0.787$\pm$0.008} \\
$\text{PPV}_{\text{SO}}$ & \scriptsize{0.125$\pm$0.217} & \scriptsize{0.417$\pm$0.333} & \scriptsize{0.119$\pm$0.168} & \scriptsize{0.094$\pm$0.011} & \scriptsize{0.155$\pm$0.139} & \scriptsize{0.282$\pm$0.058} \\
$\text{PPV}_{\text{IV}}$ & \scriptsize{0.217$\pm$0.205} & \scriptsize{0.407$\pm$0.259} & \scriptsize{0.380$\pm$0.055} & \scriptsize{0.357$\pm$0.024} & \scriptsize{0.329$\pm$0.077} & \scriptsize{0.360$\pm$0.020} \\
    \bottomrule
  \end{tabular}

\end{table}

\subsubsection{Effects of the Counterparts Selection Stringency.}
\label{appendix:counterpart-selection-study}
\begin{figure}[t]
    \centering
    \includegraphics[width=1.0\textwidth]{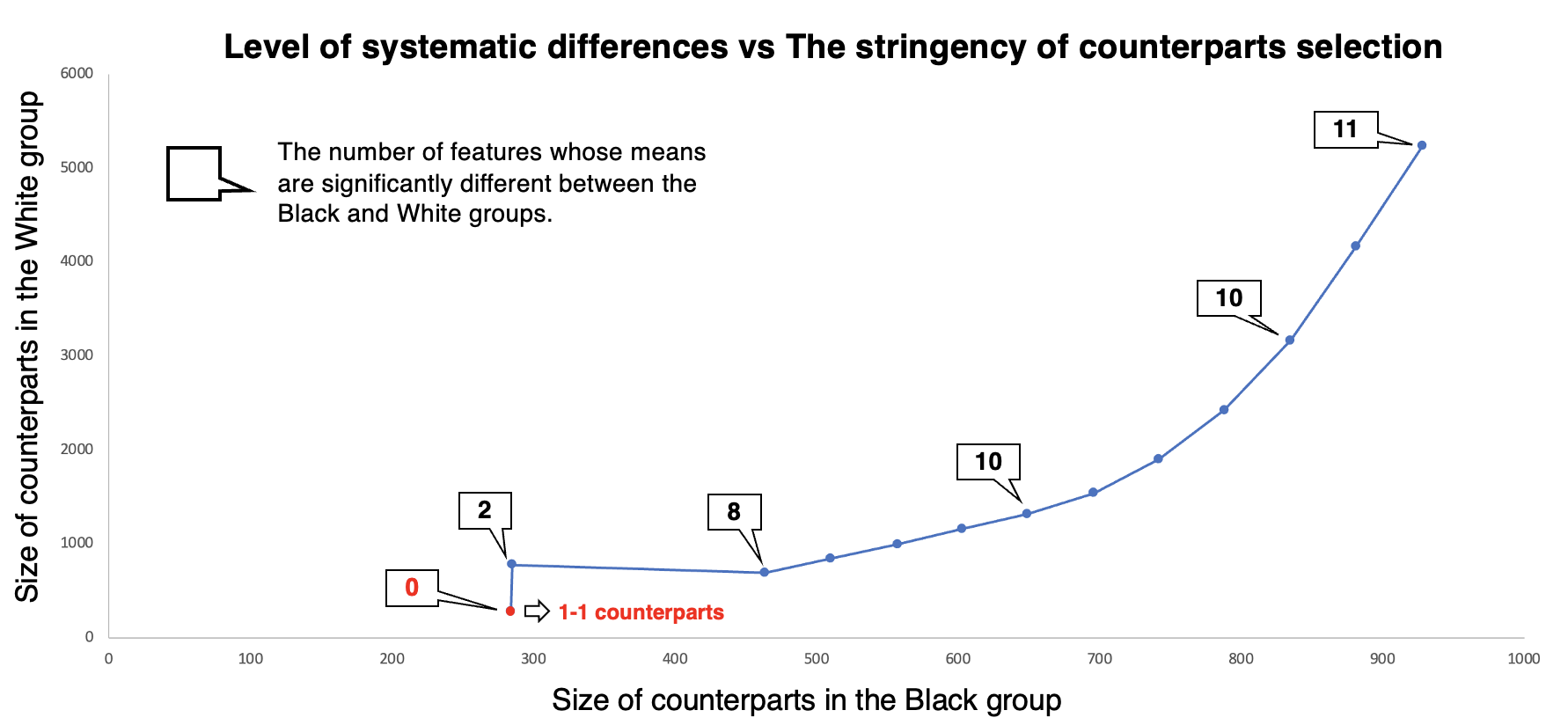}
    \caption{\small{Changes of systematic difference levels with respect to the stringency of counterpart selection in the MIMIC experiment. Loosening the counterpart similarity constraint leads to larger counterpart groups, however, increases systematic differences indicated by the increasing number of features whose means are significantly different between groups (evaluated by $t$-test, $p$-value significant level at 0.05).}}
    \label{fig:sample-selection-mimic}
\end{figure}

Fig.~\ref{fig:sample-selection-mimic} shows that reducing the stringency in selecting counterparts allows more counterparts to be identified. However, the gain in the size of counterparts can exacerbate systematic differences, evidenced by more features showing statistically significant variations in means across the selected counterpart groups. When we tightened the constraint to identify 1-1 counterparts, no features exhibit significant differences between groups, a sign of successful mitigation of systematic differences.

\subsubsection{Ablation Study of the Components in CFair Implementation}
\label{sec:appendix-ablation-study}
The current implementation of CFair mainly consists of two components: propensity score matching (PSM) and Mahalanobis distance (MD) for measuring similarity between individuals. As discussed in Section~\ref{sec:cfair-method}, PSM is used to address the issue of confounding, and the learned MD is used to guarantee that individuals in each counterpart pair are similar in their baseline characteristics. We conducted an ablation study on the MIMIC dataset to demonstrate the contributions of PSM and MD. Specifically, we examined the feature-wise differences between groups: (1) the original groups defined by protected variables, (2) the 1-1 counterpart groups selected by PSM only, and (3) the 1-1 counterpart groups chosen by combining PSM and MD. The results are summarized in Table~\ref{tab:psm-ablation}. Out of the total 16 features, 10 features exhibit statistically significant differences ($p$-value cut-off 0.05) between the original racial groups. Only 5 features show statistically significant differences between the counterpart groups decided by PSM without MD. No features show statistically significant differences between the counterpart groups if both PSM and MD are applied. The ablation study indicates the necessity of employing MD in selecting counterparts.

\begin{table}[h]
\centering
\caption{The normalized absolute mean differences between counterparts selected by three different ways on the MIMIC dataset: the original groups simply defined by the protected variable, the counterpart groups selected by PSM only, and the counterpart groups selected by PSM+MD. The statistical significance of each difference is evaluated by the $t$-test with the null hypothesis that the feature means of two group under comparison are the same. A smaller $p$-value indicates that the difference is statistically more significant. $P$-values with a significant level of 0.05 are in bold.}
\small
\begin{tabular}{l|lr|lr|lr}
\toprule
 & \multicolumn{2}{c|}{Original  groups} & \multicolumn{2}{c|}{PSM only} & \multicolumn{2}{c}{PSM + MD} \\
 \midrule
Feature & Diff. & $p$-value & Diff. & $p$-value & Diff. & $p$-value \\
 \midrule
GCS & 0.0007 & 0.073 & 0.2667 & \textbf{0.034} & 0.0002 & 0.922  \\
Sofa 24hours & 0.0820 & \textbf{0.015} & 0.8140 & $<$\textbf{0.001} & 0.0274 & 0.599 \\
HR & 0.0182 & \textbf{0.004} & 2.1146 & 0.106 & 0.0046 & 0.720 \\
SBP & 0.0002 & 0.976 & 1.7029 & 0.324 & 0.0009 & 0.064 \\
DBP & 0.0074 & 0.317 & 0.3784 & 0.692 & 0.0060 & 0.666 \\
MBP & 0.0287 & $<$\textbf{0.001} & 0.3578 & 0.747 & 0.0033 & 0.800 \\
RR & 0.0193 & \textbf{0.017} & 0.3697 & 0.349 & 0.0005 & 0.975 \\
Temperature & 0.0006 & 0.216 & 0.0948 & \textbf{0.019} & 0.0003 & 0.741 \\
Spo2 & 0.0042 & $<$\textbf{0.001} & 0.6346 & \textbf{0.012} & 0.0000 & 0.992 \\
Glucose & 0.0062 & 0.554 & 7.3556 & 0.079 & 0.0120 & 0.542 \\
Age & 0.0538 & $<$\textbf{0.001}  & 0.3895 & 0.375 & 0.0142 & 0.361 \\
CCI & 0.0677 & $<$\textbf{0.001} & 0.3123 & 0.181 & 0.0094 & 0.786 \\
APSiii & 0.0635 & $<$\textbf{0.001} & 2.1754 & 0.171 & 0.0080 & 0.759 \\
BMI & 0.0038 & 0.462  & 0.3681 & 0.732 & 0.0027 & 0.734 \\
Gender & 0.3285 & $<$\textbf{0.001} & 0.0877 & \textbf{0.034} & 0.0000 & 1.000 \\
RRT & 1.0101 & $<$\textbf{0.001} & 0.0035 & 0.862 & 0.0000 & 1.000 \\ \bottomrule
\end{tabular}
\label{tab:psm-ablation}
\end{table}

\subsection{Additional Experimental Results on the Adult and German Banking Datasets}
\label{sec:appendix-adult-gb}
\begin{figure}[t]
    \centering
    \includegraphics[width=0.8\textwidth]{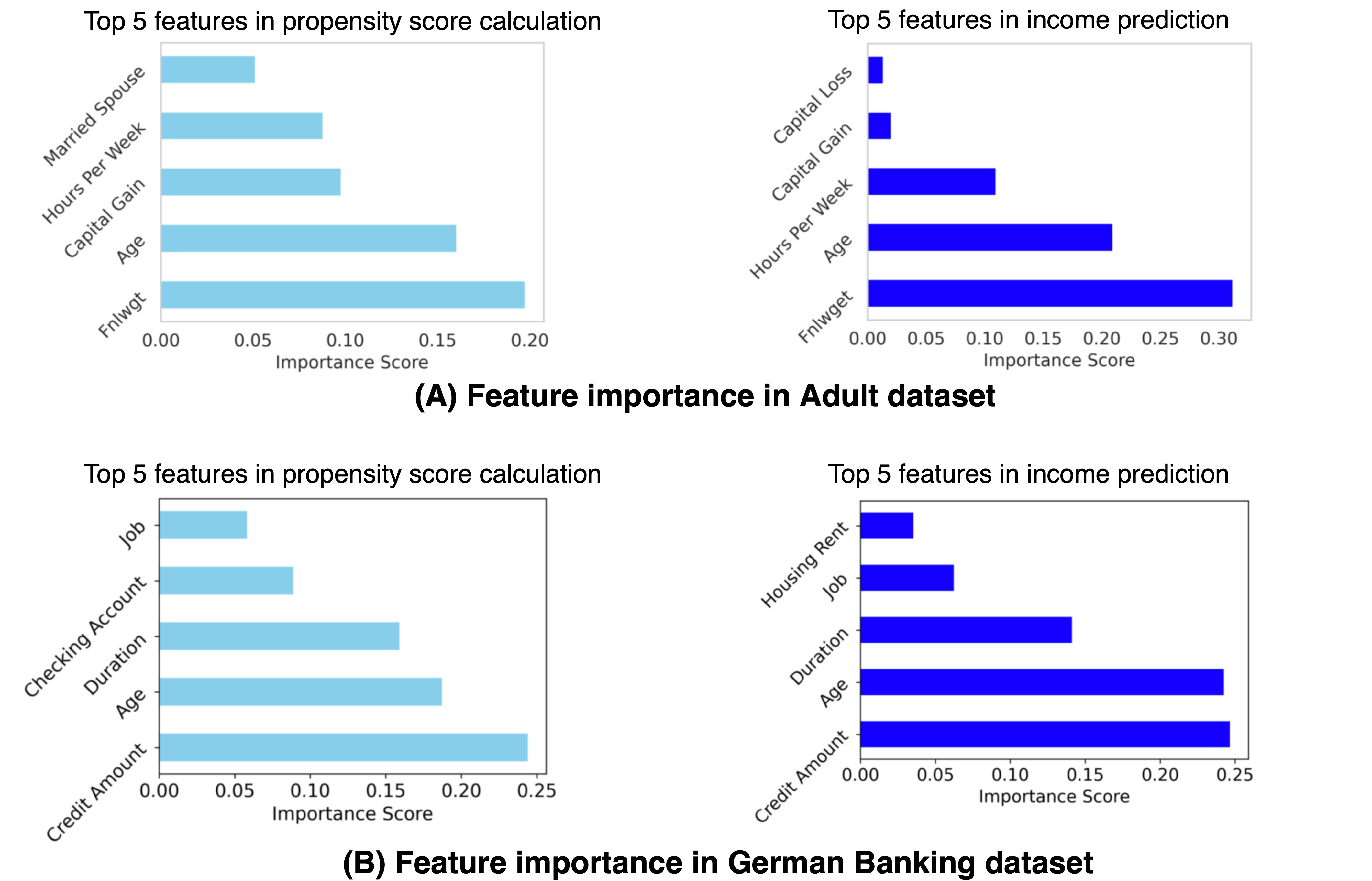}


\caption{Feature importance for propensity score calculation and income classification in Adult and German Banking dataset. (A) For Adult dataset, fnlwgt, age, working hours per week and capital gain contribute heavily to both propensity score and income prediction, which matches the common sense. (B) For German Banking dataset, credit amount, age, duration of a loan, and job contribute heavily to both propensity score and risk prediction, which matches the common sense.}
    \label{fig:app_feature-importance}
\end{figure}

Fig~\ref{fig:ps-bw} shows that the Adult and German Banking datasets contain significant systematic differences between groups. In both cases, the protected variables can be perfectly predicted by non-protected variables. That is there exist substantial confounding issues. Some non-protected variables are quite informative for both the protected variables and the target variables. For example, in the Adult dataset, there is a clear separation in propensity score distributions between the Black and White groups. This distinct separation can be traced back to several features (see Fig. \ref{fig:app_feature-importance} (A)), some of which are also important for accurately predicting income. For example, it is common sense that capital gain, age, hours per week (number of working hours per week) are influential factors in income determination. Black and White individuals exhibit distinct distributions on these features. Such systemic differences are well known to be historical in our social systems.

Similarly, Fig.~\ref{fig:app_feature-importance} (B) shows that, in the German Banking dataset, a few features (e.g., Job and Credit Amount) are excellent predictors of both the protected variable and the outcome (i.e., customer risk). It is common sense that both job and credit amount are important factors in deciding the risk class of a customer. The distributions of these two features in the female group are quite different from those in the male group, which is most likely due to the historical issues with social structures.

\subsection{Additional Results of the COMPAS Experiment}
\label{appendix:extra-disscuss-compas}
In the COMPAS dataset, most of the individuals in the Black group are not comparable to those in the White group, which is evidenced by their propensity score distributions (see gray bars in Fig.~\ref{fig:ps-bw} (B) and the significant feature-wise between-group differences (see Table~\ref{tab:t-tests-results}). This poses challenges in using traditional fairness indexes to faithfully evaluate algorithmic bias. In contrast, the counterparts identified by CFair are highly similar as indicated by the small between-group feature differences (larger $t$-test $p$-values). In other words, CFair effectively mitigates systematic differences when selecting individuals to be compared. CFair analysis (measured by either Equal Opportunity or Sufficiency) reveals that the model is significantly biased on counterparts (see Table~\ref{tab:compas-detail}). Such signals are diluted if the fairness analysis is performed on the total population.


\begin{table}[t]
  \caption{CFair analysis in the COMPAS experiment. $\Delta{\text{TPR}}$ and  $\Delta{\text{PPV}}$ represents the absolute mean difference of true positive rate (TPR) and the absolute mean difference of positive predictive value (PPV). Class 1 denotes cases where re-arrest has occurred, while Class 0 signifies otherwise.}
  \scriptsize
    \label{tab:compas-detail}
  \centering
  \begin{tabular}{l|cc|cc|cc}
    \toprule
    &  \multicolumn{2}{c}{Counterparts} & \multicolumn{2}{c}{Unmatched population} & \multicolumn{2}{c}{Total population}   
    \\
   &  \scriptsize{Black} & \scriptsize{White} &  \scriptsize{Black} & \scriptsize{White} &  \scriptsize{Black} & \scriptsize{White} \\
    \midrule
    \textbf{DP gap} &  \multicolumn{2}{c|}{0.442$\pm$0.106} & \multicolumn{2}{c|}{0.218$\pm$0.028} & \multicolumn{2}{c}{0.275$\pm$0.025} \\
    \midrule
\textbf{Accuracy} & \scriptsize{0.670$\pm$0.079} & \scriptsize{0.876$\pm$0.039} & \scriptsize{0.624$\pm$0.011} & \scriptsize{0.672$\pm$0.013} & \scriptsize{0.629$\pm$0.017} & \scriptsize{0.701$\pm$0.009} \\
\midrule
\scriptsize{\textbf{Equal Opportunity}}  & \multicolumn{2}{c|}{0.435$\pm$0.119} & \multicolumn{2}{c|}{0.191$\pm$0.026} & \multicolumn{2}{c}{0.141$\pm$0.029}  \\ 
$\Delta{\text{TPR}}$ of no re-arrest &  \multicolumn{2}{c|}{0.114$\pm$0.069} & \multicolumn{2}{c|}{0.145$\pm$0.032} & \multicolumn{2}{c}{0.146$\pm$0.037} \\
$\Delta{\text{TPR}}$ of re-arrest &  \multicolumn{2}{c|}{0.435$\pm$0.119} & \multicolumn{2}{c|}{0.191$\pm$0.026} & \multicolumn{2}{c}{0.141$\pm$0.029} \\
\midrule
\scriptsize{\textbf{Sufficiency}}  & \multicolumn{2}{c|}{0.366$\pm$0.176} & \multicolumn{2}{c|}{0.085$\pm$0.053} & \multicolumn{2}{c}{0.086$\pm$0.043}  \\
$\Delta{\text{PPV}}$ of no re-arrest &  \multicolumn{2}{c|}{0.170$\pm$0.104} & \multicolumn{2}{c|}{0.050$\pm$0.024} & \multicolumn{2}{c}{0.070$\pm$0.030} \\
$\Delta{\text{PPV}}$ of re-arrest &  \multicolumn{2}{c|}{0.337$\pm$0.187} & \multicolumn{2}{c|}{0.084$\pm$0.054} & \multicolumn{2}{c}{0.062$\pm$0.050} \\
\midrule
TPR of no re-arrest & \scriptsize{0.803$\pm$0.078} & \scriptsize{0.917$\pm$0.049} & \scriptsize{0.632$\pm$0.022} & \scriptsize{0.654$\pm$0.027} & \scriptsize{0.654$\pm$0.027} & \scriptsize{0.800$\pm$0.012} \\
TPR of re-arrest & \scriptsize{0.266$\pm$0.051} & \scriptsize{0.701$\pm$0.114} & \scriptsize{0.612$\pm$0.008} & \scriptsize{0.421$\pm$0.022} & \scriptsize{0.589$\pm$0.010} & \scriptsize{0.448$\pm$0.025} \\
\midrule
PPV of no re-arrest & \scriptsize{0.760$\pm$0.081} & \scriptsize{0.930$\pm$0.027} & \scriptsize{0.712$\pm$0.013} & \scriptsize{0.763$\pm$0.021} & \scriptsize{0.719$\pm$0.018} & \scriptsize{0.790$\pm$0.021} \\
PPV of re-arrest & \scriptsize{0.342$\pm$0.081} & \scriptsize{0.680$\pm$0.145} & \scriptsize{0.524$\pm$0.025} & \scriptsize{0.421$\pm$0.022} & \scriptsize{0.514$\pm$0.029} & \scriptsize{0.464$\pm$0.035} \\
    \bottomrule
  \end{tabular}
\end{table}

\section{Implementation Details of CFair}
\label{appendix:build PS models}
The computational resources required for our workflow involved several distinct stages. The initial candidate filtering, which employed propensity score matching on datasets such as MIMIC and COMPAS, took approximately 5 hours to complete one experiment. We use a laptop like MacBook Air (M2 chip, 16GB memory) to run it. 
The refinement step, utilizing the Mahalanobis distance metric, was more computationally intensive and took 3 hours for training on a single GPU like GeForce RTX 3090 24GB. Finally, the prediction and evaluation tasks, which involved applying the machine learning models and evaluating their performance, required about 1 hour on a local laptop. 
\subsection{Building Propensity Score Models}
Using the sepsis patients in the MIMIC dataset as an example, we elaborate on how we trained a $PS(\cdot)$ model and performed PSM. We first applied standard normalization on the input features listed in Table~\ref {tab:sepsis-feature}. This procedure can ensure equal treatment of each potential confounder and prevent one variable from dominating the analysis simply due to its scale. We tried several ML models, including logistic regression, support vector machine, decision tree, multi-layer perception and AdaBoost using trees as the base learner. The results and the hyper-parameters of the models for the Black vs White case are listed in Table~\ref{table:psm-Black-vs-White-mimic}. We report the results of random forest for Black vs White in the COMPAS experiment as shown in Table~\ref{table:psm-Black-vs-White-compas}, results for Black vs White in the Adult experiment as shown in Table~\ref{table:psm-Black-vs-White-adult}, as well as results for Female vs Male in the German Banking experiment as shown in Table~\ref{table:psm-female-vs-male-gb}.

\begin{table}[t]
  \caption{Propensity score models (Black vs White) in the MIMIC experiment. MLP stands for Multi-Layer Perceptron}
  \label{table:psm-Black-vs-White-mimic}
  \centering
  \begin{tabular}{lcccl}
    \toprule
    \scriptsize{Model}   &\scriptsize{F1 on the Black group} \ \ & \scriptsize{F1 on the White group}   &\quad \ \  \scriptsize{Macro F1} \ \qquad \quad& \scriptsize{\ \ \ Best hyperparameters}  \\
    \midrule
     \scriptsize{Logistic Regression} &\hspace{-1em} \scriptsize{0.75} &\hspace{-1.5em} \scriptsize{0.26} & \scriptsize{0.63} & \begin{tabular}{l} \scriptsize{Penalty: L2} \\  \scriptsize{Class weight: balanced} \end{tabular}\\
    \midrule
    \scriptsize{Adaboost} &\hspace{-1em} \scriptsize{\textbf{0.97}} & \hspace{-1em}\scriptsize{\textbf{0.80}} & \scriptsize{\textbf{0.89}} & \begin{tabular}{l} \scriptsize{Class weight: balanced} \\
    \scriptsize{Number of estimators: 200}
    \end{tabular} \\
    \midrule
    \scriptsize{Random Forest} &\hspace{-1em} \scriptsize{0.96} & \hspace{-1.2em}\scriptsize{0.71} & \scriptsize{0.83} & \begin{tabular}{l}
    \scriptsize{Max depth: 10} \\
    \scriptsize{Number of estimators: 200} \\
    \end{tabular} \\
    \midrule
    \scriptsize{MLP}  &\hspace{-1em} \scriptsize{0.94} &\hspace{-1.4em} \scriptsize{0.00} & \scriptsize{0.47} & \begin{tabular}{l}
    \scriptsize{1 hidden layer with 20 neurons} \\
    \scriptsize{Learning rate: 0.05} \\
   \end{tabular} \\
    \bottomrule
  \end{tabular}
\end{table}

\begin{table}[h]
  \caption{Propensity score models (Black vs White) in the COMPAS experiment}
  \label{table:psm-Black-vs-White-compas}
  \centering
  \begin{tabular}{lcccl}
    \toprule
    \scriptsize{Model}   &\scriptsize{F1 on the Black group} \ \ & \scriptsize{F1 on the White group}   &\quad \ \  \scriptsize{Macro F1} \ \qquad \quad& \scriptsize{\ \ \ Best hyperparameters}  \\
    \hline
    \scriptsize{Logistic Regression}   &\hspace{-1em}  \scriptsize{0.60}   &\hspace{-1.5em}  \scriptsize{0.67}    &  \scriptsize{0.63} & \begin{tabular}{l}  \scriptsize{Penalty: L2} \\  \scriptsize{Class weight: balanced} \end{tabular}  \\
    \hline
    \scriptsize{AdaBoost}    &\hspace{-1em} \scriptsize{0.96}   &\hspace{-1.5em} \scriptsize{\textbf{0.98}}    &  \scriptsize{\textbf{0.97}} & \begin{tabular}{l} \scriptsize{Number of estimators: 100} \\  \scriptsize{Base estimator: decision tree} \end{tabular}  \\
    \hline
    \scriptsize{Random Forest}    &\hspace{-1em} \scriptsize{\textbf{0.98}}   &\hspace{-1.5em} \scriptsize{0.97}    & \scriptsize{\textbf{0.97}} & \begin{tabular}{l}  \scriptsize{Number of estimators: 200} \\  \scriptsize{Base estimator: decision tree} \end{tabular}  \\
    \bottomrule
  \end{tabular}
  
\end{table}

\begin{table}[h!]
  \caption{Propensity score model (female vs male) in the German Banking experiment}
  \label{table:psm-female-vs-male-gb}
  \centering
  \begin{tabular}{lcccl}
    \toprule
    \scriptsize{Model}    &\quad \scriptsize{\quad F1 on the female group} & \scriptsize{F1 on the male group}  &\qquad  \scriptsize{Macro F1} \ \qquad \quad& \scriptsize{\ \ \ Best hyperparameters}  \\
    \hline
    \scriptsize{Random Forest}    &\quad \scriptsize{\textbf{1.00}}   &\hspace{-1em}\scriptsize{\textbf{1.00}}    & \scriptsize{\textbf{1.00}} & \begin{tabular}{l}  \scriptsize{Number of estimators: 100} \\  \scriptsize{Base estimator: decision tree}  \end{tabular}  \\
    \bottomrule
  \end{tabular}
  
\end{table}

\begin{table}[b]
  \caption{Propensity score model (Black vs White) in the Adult experiment}
  \label{table:psm-Black-vs-White-adult}
  \centering
  \begin{tabular}{lcccl}
    \toprule
    \scriptsize{Model}    &\quad \scriptsize{\quad F1 on the Black group} \ \ & \scriptsize{F1 on the White group}  &\quad \ \  \scriptsize{Macro F1} \ \qquad \quad & \scriptsize{\ \ \ Best hyperparameters}  \\
    \hline
    \scriptsize{Random Forest}    &\quad \scriptsize{\textbf{1.00}}   &\hspace{-1.5em}\scriptsize{\textbf{1.00}}    & \hspace{-0.5em}\scriptsize{\textbf{1.00}} & \begin{tabular}{l}  \scriptsize{Number of estimators: 100} \\  \scriptsize{Base estimator: decision tree}  \end{tabular}  \\
    \bottomrule
  \end{tabular}
  
\end{table}

\subsection{Learning the Distance Metric}
\label{appendix:cov-learner}

We applied gradient descent to find $\boldsymbol{W}$ of the Mahalanobis distance eq.\ref{equ:dissimilarity-function} by optimizing the cost function eq. (\ref{eq:cost-function}). $\boldsymbol{W}$ was first initialized using the inverse of the weighted covariance matrix, which was determined by taking the weighted sum of the covariance matrices of the samples chosen by PSM. The gradient descent procedure used a learning rate of 0.0001 and set the maximal iteration to 100. 
\begin{table}[hb!]
  \caption{The ventilation prediction performance in the MIMIC experiment.}
  \label{table:vent-Black-White}
  \small
  \centering
  \begin{tabular}{lcccc}
    \toprule
    \normalsize{F1 Scores} & Random Forest & Logistic Regression & AdaBoost \\
    \midrule
    No Ventilation & \textbf{0.834$\pm$0.005} & 0.494$\pm$0.016 & 0.774$\pm$0.005 \\
    Supplement Oxygen & 0.128$\pm$0.044 & \textbf{0.268$\pm$0.013} & 0.182$\pm$0.007 \\
    Invasive Ventilation & 0.306$\pm$0.017 & \textbf{0.320$\pm$0.013} & 0.194$\pm$0.034 \\
    Macro F1 & \textbf{0.424$\pm$0.014} & 0.360$\pm$0.011 & 0.382$\pm$0.016 \\
    \bottomrule
  \end{tabular}
\end{table}

\begin{table}[hb!]
  \caption{The recidivism prediction performance in the COMPAS experiment.}
  \label{tab:recid-Black-White}
  \small
  \centering
  \begin{tabular}{lcccc}
    \toprule
    \normalsize{F1 Scores} & Random Forest & Logistic Regression & AdaBoost \\
    \midrule
    No recidivism & \textbf{0.804$\pm$0.007} & 0.709$\pm$0.009 & 0.714$\pm$0.009 \\
    Recidivism & \textbf{0.598$\pm$0.015} & 0.526$\pm$0.012 & 0.527$\pm$0.010 \\
    Macro F1 & \textbf{0.721$\pm$0.038} & 0.618$\pm$0.007 & 0.620$\pm$0.002 \\
    \bottomrule
  \end{tabular}
\end{table}

\subsection{Training Ventilation Prediction Models in the MIMIC Experiment}
\label{appendix:vent-training}
There are three ventilation statuses: (0) No-ventilation, (1) Supplemental oxygen, (2) and Invasive ventilation. We tested several machine learning techniques for predicting ventilation status (results in Table~\ref{table:vent-Black-White}). Cross entropy was used as the loss function. Five-fold cross-validation was used to tune the hyper-parameters of each model. It was made sure that both counterparts and non-counterparts were randomly split in the same way in each cross-validation run. To address the data imbalance issue, we applied the SMOTE technique \cite{chawla2002smote} to augment data of the minority classes (i.e., Types 1 and 2) during training. The results are summarized in Table~\ref{table:vent-Black-White}. Finally, we chose random forest as our ventilation prediction model.

\subsection{Training Recidivism Prediction Models in the COMPAS Experiment}
\label{appendix:recid-training}
The procedure for training the recidivism prediction model is similar to the one used in the MIMIC experiment. The results are summarized in Table~\ref{tab:recid-Black-White}.

\newpage
\section{CFair Analysis on Synthetic Datasets}
\label{appendix:synthetic_experiment}
We conducted the following experiment using synthetic datasets, in which the counterpart ground truth is known, to demonstrate how to use CFair to reveal different aspects of algorithmic fairness. DP gap and CDP gap were used in this experiment. This experiment emphasizes algorithmic fairness analysis.

\begin{figure}[h]
    \centering
    \includegraphics[width=0.9\textwidth]{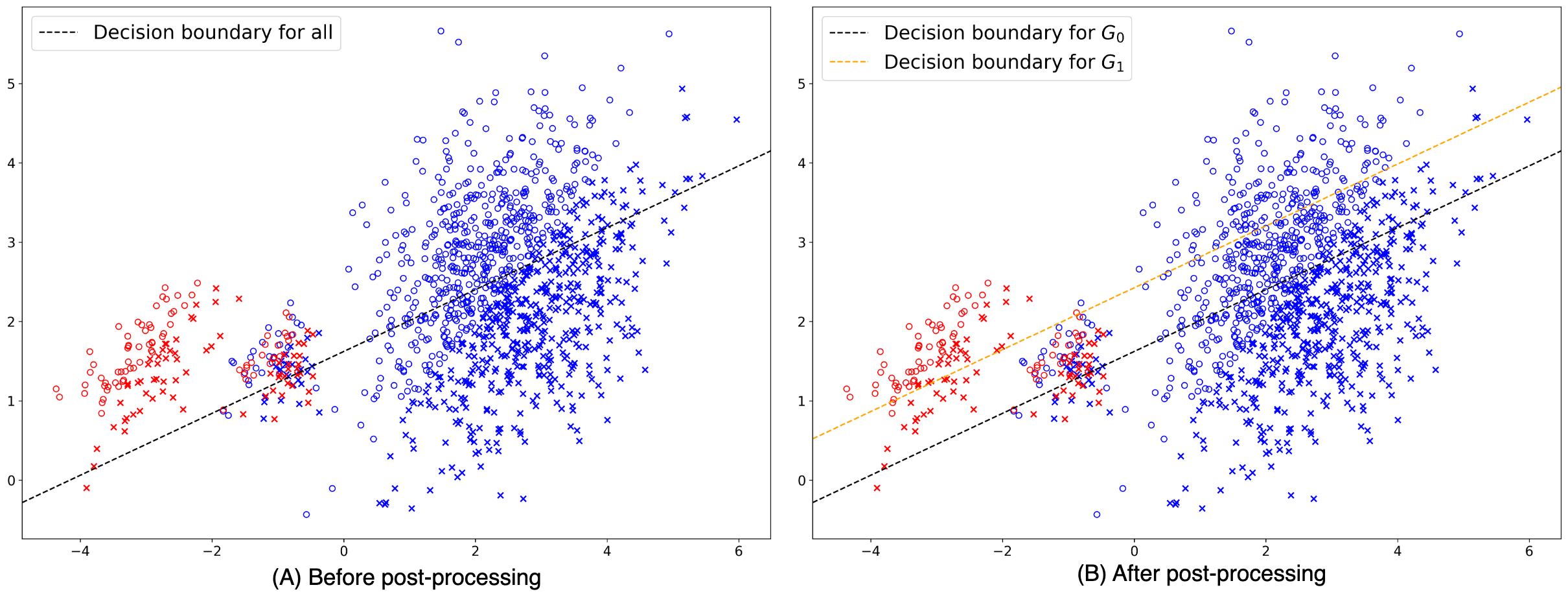}
    \caption{\small{Experiment using synthetic data. (a) There are two groups ($G_0$ red vs $G_1$ blue) and two classes (`o' vs `x'). The distributions of $G_0$ and $G_1$ exhibit obvious systematic differences. Counterparts can be found in the region where they overlap. A logistic regression model is fit to classify samples in to `o' or `x' classes. The decision boundary of the model is shown as the dash line. Since $G_1$ dominates the data, the model is trained to favor $G_1$ and perform substantially worse on $G_0$. (B) To increase the accuracy of the model on $G_0$, a post-processing step is applied to adjust the threshold of the model for the $G_0$ samples, producing the red dash line decision boundary.}}
    \label{fig:synthesis-data}
\end{figure}
\begin{table}[h]
  \caption{DP gap and CDP gap before and after post-processing. Each experiment is repeated 100 times, and the average value and standard deviation are reported.}
\label{table:synthetic-bias}
  \small
  \centering
  \begin{tabular}{lcc}
    \toprule
     & DP gap & CDP gap\\
        \midrule
   Before post-processing  & 0.445 $\pm$ 0.041 & 0.038 $\pm$ 0.028\\
   \midrule
   After post-processing & 0.065 $\pm$ 0.047 & 0.708 $\pm$ 0.097  \\
    \bottomrule
  \end{tabular}
  \\
\end{table}

\paragraph{Simulation.} 
We simulate $G_0$ from a mixture of two Gaussian components ($\mathcal{N}_0^{1}$ and $\mathcal{N}^{shared}$): $\mathcal{N}_0^{1}$ has a mean of (-3, 1.5) and a covariance of $\begin{bmatrix}0.3 & 0.2\\0.2 & 0.3\end{bmatrix}$, and $\mathcal{N}^{shared}$ has a mean (-1, 1.5) and a covariance of $\begin{bmatrix}0.1 & 0.05\\0.05 & 0.1\end{bmatrix}$. The $G_1$ samples are simulated from another mixture of two Gaussian components ($\mathcal{N}_1^{1}$ and $\mathcal{N}^{shared}$): $\mathcal{N}_1^{1}$ has a mean of (2,5, 2.5) and a covariance of $\begin{bmatrix}1 & 0.3\\0.3 & 1\end{bmatrix}$, and $\mathcal{N}^{shared}$. $\mathcal{N}^{shared}$ represent where two groups overlap, and $\mathcal{N}_0^{1}$ and $\mathcal{N}_1^{1}$ indicate where systematic differences between $G_0$ and $G_1$ exist. One hundred samples in $G_0$ are sampled from $\mathcal{N}_0^{1}$, and 1,000 samples in $G_1$ are from $\mathcal{N}_1^{1}$. These 1,100 samples represent the systematic differences between $G_0$ and $G_1$. We generate 50 counterpart pairs by sampling 50 samples for $G_1$ from $\mathcal{N}^{shared}$ and add small amount of Gaussian noise $\mathcal{N}(0, 0.01*I)$ to those samples to produce their counterparts in $G_0$. Note: For the purpose of clear illustration, we opt to generate data in 2D space and deliberately design $\mathcal{N}_0^{1}$, $\mathcal{N}_1^{1}$ and $\mathcal{N}^{shared}$ to be well separated.




 Class labels are assigned as the following. Samples from $\mathcal{N}_0^{1}$ are assigned to class `1' if they satisfy \(x_2 - x_1 - 4.5 > 0\), and `0' otherwise. Samples from $\mathcal{N}^{shared}$ are assigned to class `1' if they satisfy \(x_2 - x_1 - 2.5 > 0\), and `0' otherwise.The rest of the samples are assigned to class `1' if they satisfy \(x_2 - x_1 > 0\), and `0' otherwise. The above procedure labels counterparts in the same way, however, treats non-counterpart samples in $G_0$ differently, contributing to systematic differences.

\paragraph{Classifier Construction.}
We trained a logistic regression model (see the black dashed line in Fig. \ref{fig:synthesis-data}A) using all samples. This model misclassifies a large number of $G_0$ samples simulated from $\mathcal{N}_0^{1}$, resulting in a large DP gap (0.445 $\pm$ 0.041). This is not of surprise because the samples from the $G_1$ group dominate the dataset. However, this model treats counterparts fairly as it produces a small CDP gap (0.038 $\pm$ 0.028).

To reduce the DP gap of the above model, a simple post-process step \cite{hort2023bias} can be adopted by adjusting the decision threshold from 0.5 to 0.85 for $G_0$ samples (the orange dashed line in Fig \ref{fig:synthesis-data}B). This adjustment significantly improved the accuracy on $G_0$, and effectively reduced its DP gap value to 0.065 $\pm$ 0.047. However, this modification results in the reclassification of nearly all positive counterparts in $G_0$, leading to a huge CDP gap (0.708 $\pm$ 0.097).

\paragraph{Discussion.} This experiment shows that CFair analysis allows users to reveal algorithmic unfairness issues, which may not be detected by traditional fairness analysis. It also suggested that, for the purpose of increasing both model performance and algorithmic fairness, it may be better to identify regions where systematic differences exist and build a model for each region. For example, in the context of this experiment, one may want to build a model for samples from $\mathcal{N}_0^{1}$ and another model for samples from $\mathcal{N}_1^{1}$ and $\mathcal{N}^{shared}$. In real applications, it might not always be feasible to develop a model tailored for every individual region. Nonetheless, it remains valuable to pinpoint areas where a model exhibits bias.

\newpage
\section*{NeurIPS Paper Checklist}


\begin{enumerate}

\item {\bf Claims}
    \item[] Question: Do the main claims made in the abstract and introduction accurately reflect the paper's contributions and scope?
    \item[] Answer: \answerYes{}  
    \item[] Justification: The abstract and introduction highlights that the paper's scope is to propose a more refined and comprehensive fairness index that counts for systematic differences.
    \item[] Guidelines:
    \begin{itemize}
        \item The answer NA means that the abstract and introduction do not include the claims made in the paper.
        \item The abstract and/or introduction should clearly state the claims made, including the contributions made in the paper and important assumptions and limitations. A No or NA answer to this question will not be perceived well by the reviewers. 
        \item The claims made should match theoretical and experimental results, and reflect how much the results can be expected to generalize to other settings. 
        \item It is fine to include aspirational goals as motivation as long as it is clear that these goals are not attained by the paper. 
    \end{itemize}

\item {\bf Limitations}
    \item[] Question: Does the paper discuss the limitations of the work performed by the authors?
    \item[] Answer: \answerYes{} 
    \item[] Justification: We discussed the limitations and future work in Appendix~\ref{appendix:future}.
    \item[] Guidelines:
    \begin{itemize}
        \item The answer NA means that the paper has no limitation while the answer No means that the paper has limitations, but those are not discussed in the paper. 
        \item The authors are encouraged to create a separate "Limitations" section in their paper.
        \item The paper should point out any strong assumptions and how robust the results are to violations of these assumptions (e.g., independence assumptions, noiseless settings, model well-specification, asymptotic approximations only holding locally). The authors should reflect on how these assumptions might be violated in practice and what the implications would be.
        \item The authors should reflect on the scope of the claims made, e.g., if the approach was only tested on a few datasets or with a few runs. In general, empirical results often depend on implicit assumptions, which should be articulated.
        \item The authors should reflect on the factors that influence the performance of the approach. For example, a facial recognition algorithm may perform poorly when image resolution is low or images are taken in low lighting. Or a speech-to-text system might not be used reliably to provide closed captions for online lectures because it fails to handle technical jargon.
        \item The authors should discuss the computational efficiency of the proposed algorithms and how they scale with dataset size.
        \item If applicable, the authors should discuss possible limitations of their approach to address problems of privacy and fairness.
        \item While the authors might fear that complete honesty about limitations might be used by reviewers as grounds for rejection, a worse outcome might be that reviewers discover limitations that aren't acknowledged in the paper. The authors should use their best judgment and recognize that individual actions in favor of transparency play an important role in developing norms that preserve the integrity of the community. Reviewers will be specifically instructed to not penalize honesty concerning limitations.
    \end{itemize}

\item {\bf Theory Assumptions and Proofs}
    \item[] Question: For each theoretical result, does the paper provide the full set of assumptions and a complete (and correct) proof?
    \item[] Answer :\answerYes{} 
    \item[] Justification: We provided the full set of assumptions and a complete proof in Appendix~\ref{appendix:biased-sampling}.
    \item[] Guidelines:
    \begin{itemize}
        \item The answer NA means that the paper does not include theoretical results. 
        \item All the theorems, formulas, and proofs in the paper should be numbered and cross-referenced.
        \item All assumptions should be clearly stated or referenced in the statement of any theorems.
        \item The proofs can either appear in the main paper or the supplemental material, but if they appear in the supplemental material, the authors are encouraged to provide a short proof sketch to provide intuition. 
        \item Inversely, any informal proof provided in the core of the paper should be complemented by formal proofs provided in appendix or supplemental material.
        \item Theorems and Lemmas that the proof relies upon should be properly referenced. 
    \end{itemize}

    \item {\bf Experimental Result Reproducibility}
    \item[] Question: Does the paper fully disclose all the information needed to reproduce the main experimental results of the paper to the extent that it affects the main claims and/or conclusions of the paper (regardless of whether the code and data are provided or not)?
    \item[] Answer: \answerYes{} 
    \item[] Justification: We provided detailed descriptions of the experiments for reproduction.
    \item[] Guidelines:
    \begin{itemize}
        \item The answer NA means that the paper does not include experiments.
        \item If the paper includes experiments, a No answer to this question will not be perceived well by the reviewers: Making the paper reproducible is important, regardless of whether the code and data are provided or not.
        \item If the contribution is a dataset and/or model, the authors should describe the steps taken to make their results reproducible or verifiable. 
        \item Depending on the contribution, reproducibility can be accomplished in various ways. For example, if the contribution is a novel architecture, describing the architecture fully might suffice, or if the contribution is a specific model and empirical evaluation, it may be necessary to either make it possible for others to replicate the model with the same dataset, or provide access to the model. In general. releasing code and data is often one good way to accomplish this, but reproducibility can also be provided via detailed instructions for how to replicate the results, access to a hosted model (e.g., in the case of a large language model), releasing of a model checkpoint, or other means that are appropriate to the research performed.
        \item While NeurIPS does not require releasing code, the conference does require all submissions to provide some reasonable avenue for reproducibility, which may depend on the nature of the contribution. For example
        \begin{enumerate}
            \item If the contribution is primarily a new algorithm, the paper should make it clear how to reproduce that algorithm.
            \item If the contribution is primarily a new model architecture, the paper should describe the architecture clearly and fully.
            \item If the contribution is a new model (e.g., a large language model), then there should either be a way to access this model for reproducing the results or a way to reproduce the model (e.g., with an open-source dataset or instructions for how to construct the dataset).
            \item We recognize that reproducibility may be tricky in some cases, in which case authors are welcome to describe the particular way they provide for reproducibility. In the case of closed-source models, it may be that access to the model is limited in some way (e.g., to registered users), but it should be possible for other researchers to have some path to reproducing or verifying the results.
        \end{enumerate}
    \end{itemize}

\item {\bf Open access to data and code}
    \item[] Question: Does the paper provide open access to the data and code, with sufficient instructions to faithfully reproduce the main experimental results, as described in supplemental material?
    \item[] Answer: \answerYes{} 
    \item[] Justification: We provided the code in the supplementary materials.
    \item[] Guidelines:
    \begin{itemize}
        \item The answer NA means that paper does not include experiments requiring code.
        \item Please see the NeurIPS code and data submission guidelines (\url{https://nips.cc/public/guides/CodeSubmissionPolicy}) for more details.
        \item While we encourage the release of code and data, we understand that this might not be possible, so “No” is an acceptable answer. Papers cannot be rejected simply for not including code, unless this is central to the contribution (e.g., for a new open-source benchmark).
        \item The instructions should contain the exact command and environment needed to run to reproduce the results. See the NeurIPS code and data submission guidelines (\url{https://nips.cc/public/guides/CodeSubmissionPolicy}) for more details.
        \item The authors should provide instructions on data access and preparation, including how to access the raw data, preprocessed data, intermediate data, and generated data, etc.
        \item The authors should provide scripts to reproduce all experimental results for the new proposed method and baselines. If only a subset of experiments are reproducible, they should state which ones are omitted from the script and why.
        \item At submission time, to preserve anonymity, the authors should release anonymized versions (if applicable).
        \item Providing as much information as possible in supplemental material (appended to the paper) is recommended, but including URLs to data and code is permitted.
    \end{itemize}

\item {\bf Experimental Setting/Details}
    \item[] Question: Does the paper specify all the training and test details (e.g., data splits, hyperparameters, how they were chosen, type of optimizer, etc.) necessary to understand the results?
    \item[] Answer: \answerYes{} 
    \item[] Justification: We provided detailed settings of the experiments in the main text and additional explanations in the appendix.
    \item[] Guidelines: 
    \begin{itemize}
        \item The answer NA means that the paper does not include experiments.
        \item The experimental setting should be presented in the core of the paper to a level of detail that is necessary to appreciate the results and make sense of them.
        \item The full details can be provided either with the code, in appendix, or as supplemental material.
    \end{itemize}

\item {\bf Experiment Statistical Significance}
    \item[] Question: Does the paper report error bars suitably and correctly defined or other appropriate information about the statistical significance of the experiments?
    \item[] Answer: \answerYes{} 
    \item[] Justification: We reported both averaged values, standard deviations in experiments. P-values were reported in the experiment of comparing between-group results. 
    \item[] Guidelines:
    \begin{itemize}
        \item The answer NA means that the paper does not include experiments.
        \item The authors should answer "Yes" if the results are accompanied by error bars, confidence intervals, or statistical significance tests, at least for the experiments that support the main claims of the paper.
        \item The factors of variability that the error bars are capturing should be clearly stated (for example, train/test split, initialization, random drawing of some parameter, or overall run with given experimental conditions).
        \item The method for calculating the error bars should be explained (closed form formula, call to a library function, bootstrap, etc.)
        \item The assumptions made should be given (e.g., Normally distributed errors).
        \item It should be clear whether the error bar is the standard deviation or the standard error of the mean.
        \item It is OK to report 1-sigma error bars, but one should state it. The authors should preferably report a 2-sigma error bar than state that they have a 96\% CI, if the hypothesis of Normality of errors is not verified.
        \item For asymmetric distributions, the authors should be careful not to show in tables or figures symmetric error bars that would yield results that are out of range (e.g. negative error rates).
        \item If error bars are reported in tables or plots, The authors should explain in the text how they were calculated and reference the corresponding figures or tables in the text.
    \end{itemize}

\item {\bf Experiments Compute Resources}
    \item[] Question: For each experiment, does the paper provide sufficient information on the computer resources (type of compute workers, memory, time of execution) needed to reproduce the experiments?
    \item[] Answer: \answerYes{} 
    \item[] Justification: We provided the computer resources in Appendix~\ref{appendix:build PS models}.
    \item[] Guidelines:
    \begin{itemize}
        \item The answer NA means that the paper does not include experiments.
        \item The paper should indicate the type of compute workers CPU or GPU, internal cluster, or cloud provider, including relevant memory and storage.
        \item The paper should provide the amount of compute required for each of the individual experimental runs as well as estimate the total compute. 
        \item The paper should disclose whether the full research project required more compute than the experiments reported in the paper (e.g., preliminary or failed experiments that didn't make it into the paper). 
    \end{itemize}
    
\item {\bf Code Of Ethics}
    \item[] Question: Does the research conducted in the paper conform, in every respect, with the NeurIPS Code of Ethics \url{https://neurips.cc/public/EthicsGuidelines}?
    \item[] Answer: \answerYes{} 
    \item[] Justification: The research conformed with the NeurIPS Code of Ethics.
    \item[] Guidelines:
    \begin{itemize}
        \item The answer NA means that the authors have not reviewed the NeurIPS Code of Ethics.
        \item If the authors answer No, they should explain the special circumstances that require a deviation from the Code of Ethics.
        \item The authors should make sure to preserve anonymity (e.g., if there is a special consideration due to laws or regulations in their jurisdiction).
    \end{itemize}

\item {\bf Broader Impacts}
    \item[] Question: Does the paper discuss both potential positive societal impacts and negative societal impacts of the work performed?
    \item[] Answer: \answerYes{} 
    \item[] Justification: We discussed broader impacts and limitations in Appendix~\ref{appendix:future}.
    \item[] Guidelines:
    \begin{itemize}
        \item The answer NA means that there is no societal impact of the work performed.
        \item If the authors answer NA or No, they should explain why their work has no societal impact or why the paper does not address societal impact.
        \item Examples of negative societal impacts include potential malicious or unintended uses (e.g., disinformation, generating fake profiles, surveillance), fairness considerations (e.g., deployment of technologies that could make decisions that unfairly impact specific groups), privacy considerations, and security considerations.
        \item The conference expects that many papers will be foundational research and not tied to particular applications, let alone deployments. However, if there is a direct path to any negative applications, the authors should point it out. For example, it is legitimate to point out that an improvement in the quality of generative models could be used to generate deepfakes for disinformation. On the other hand, it is not needed to point out that a generic algorithm for optimizing neural networks could enable people to train models that generate Deepfakes faster.
        \item The authors should consider possible harms that could arise when the technology is being used as intended and functioning correctly, harms that could arise when the technology is being used as intended but gives incorrect results, and harms following from (intentional or unintentional) misuse of the technology.
        \item If there are negative societal impacts, the authors could also discuss possible mitigation strategies (e.g., gated release of models, providing defenses in addition to attacks, mechanisms for monitoring misuse, mechanisms to monitor how a system learns from feedback over time, improving the efficiency and accessibility of ML).
    \end{itemize}
    
\item {\bf Safeguards}
    \item[] Question: Does the paper describe safeguards that have been put in place for responsible release of data or models that have a high risk for misuse (e.g., pretrained language models, image generators, or scraped datasets)?
    \item[] Answer: \answerNA{} 
    \item[] Justification: This paper poses no such risks.
    \item[] Guidelines:
    \begin{itemize}
        \item The answer NA means that the paper poses no such risks.
        \item Released models that have a high risk for misuse or dual-use should be released with necessary safeguards to allow for controlled use of the model, for example by requiring that users adhere to usage guidelines or restrictions to access the model or implementing safety filters. 
        \item Datasets that have been scraped from the Internet could pose safety risks. The authors should describe how they avoided releasing unsafe images.
        \item We recognize that providing effective safeguards is challenging, and many papers do not require this, but we encourage authors to take this into account and make a best faith effort.
    \end{itemize}

\item {\bf Licenses for existing assets}
    \item[] Question: Are the creators or original owners of assets (e.g., code, data, models), used in the paper, properly credited and are the license and terms of use explicitly mentioned and properly respected?
    \item[] Answer: \answerYes{} 
    \item[] Justification: We cited the original paper that produced the code package or dataset in the main text.
    \item[] Guidelines:
    \begin{itemize}
        \item The answer NA means that the paper does not use existing assets.
        \item The authors should cite the original paper that produced the code package or dataset.
        \item The authors should state which version of the asset is used and, if possible, include a URL.
        \item The name of the license (e.g., CC-BY 4.0) should be included for each asset.
        \item For scraped data from a particular source (e.g., website), the copyright and terms of service of that source should be provided.
        \item If assets are released, the license, copyright information, and terms of use in the package should be provided. For popular datasets, \url{paperswithcode.com/datasets} has curated licenses for some datasets. Their licensing guide can help determine the license of a dataset.
        \item For existing datasets that are re-packaged, both the original license and the license of the derived asset (if it has changed) should be provided.
        \item If this information is not available online, the authors are encouraged to reach out to the asset's creators.
    \end{itemize}

\item {\bf New Assets}
    \item[] Question: Are new assets introduced in the paper well documented and is the documentation provided alongside the assets?
    \item[] Answer: \answerYes{} 
    \item[] Justification: We provided documentation of the code in the supplementary materials. 
    \item[] Guidelines:
    \begin{itemize}
        \item The answer NA means that the paper does not release new assets.
        \item Researchers should communicate the details of the dataset/code/model as part of their submissions via structured templates. This includes details about training, license, limitations, etc. 
        \item The paper should discuss whether and how consent was obtained from people whose asset is used.
        \item At submission time, remember to anonymize your assets (if applicable). You can either create an anonymized URL or include an anonymized zip file.
    \end{itemize}

\item {\bf Crowdsourcing and Research with Human Subjects}
    \item[] Question: For crowdsourcing experiments and research with human subjects, does the paper include the full text of instructions given to participants and screenshots, if applicable, as well as details about compensation (if any)? 
    \item[] Answer: \answerNA{} 
    \item[] Justification: This paper does not involve crowdsourcing nor research with human subjects.
    \item[] Guidelines:
    \begin{itemize}
        \item The answer NA means that the paper does not involve crowdsourcing nor research with human subjects.
        \item Including this information in the supplemental material is fine, but if the main contribution of the paper involves human subjects, then as much detail as possible should be included in the main paper. 
        \item According to the NeurIPS Code of Ethics, workers involved in data collection, curation, or other labor should be paid at least the minimum wage in the country of the data collector. 
    \end{itemize}

\item {\bf Institutional Review Board (IRB) Approvals or Equivalent for Research with Human Subjects}
    \item[] Question: Does the paper describe potential risks incurred by study participants, whether such risks were disclosed to the subjects, and whether Institutional Review Board (IRB) approvals (or an equivalent approval/review based on the requirements of your country or institution) were obtained?
    \item[] Answer: \answerNA{} 
    \item[] Justification: This paper does not involve crowdsourcing nor research with human subjects.
    \item[] Guidelines:
    \begin{itemize}
        \item The answer NA means that the paper does not involve crowdsourcing nor research with human subjects.
        \item Depending on the country in which research is conducted, IRB approval (or equivalent) may be required for any human subjects research. If you obtained IRB approval, you should clearly state this in the paper. 
        \item We recognize that the procedures for this may vary significantly between institutions and locations, and we expect authors to adhere to the NeurIPS Code of Ethics and the guidelines for their institution. 
        \item For initial submissions, do not include any information that would break anonymity (if applicable), such as the institution conducting the review.
    \end{itemize}

\end{enumerate}

\end{document}